\documentclass[10pt,twocolumn,letterpaper]{article}

\usepackage{cvpr}              

\PassOptionsToPackage{pagebackref=false}{hyperref}

\definecolor{cvprblue}{rgb}{0.21,0.49,0.74}
\usepackage[breaklinks,colorlinks,allcolors=cvprblue]{hyperref}

\usepackage{algorithm}
\usepackage{algorithmic}

\usepackage{adjustbox}
\usepackage{multirow}

\usepackage{pifont}
\usepackage{color, colortbl}
\usepackage{amsfonts} 
\usepackage{comment}
\usepackage{subcaption}
\usepackage{array}
\usepackage{tikz} 
\usepackage{acronym}
\usepackage{xspace}
\usepackage{cuted}

\definecolor{Color1}{gray}{0.95}
\definecolor{Color2}{gray}{0.87}
\definecolor{methodcolor}{HTML}{8CE4FF}
\definecolor{mygreen}{rgb}{0.032, 0.6392, 0.2039}
\definecolor{nmgray}{RGB}{229,229,229}
\definecolor{mycolor}{HTML}{FF5F00}
\definecolor{delta_color}{HTML}{16C47F}
\definecolor{kangColor}{rgb}{1,0.33,0.64}

\makeatletter
\DeclareRobustCommand\onedot{\futurelet\@let@token\@onedot}
\def\@onedot{\ifx\@let@token.\else.\null\fi}

\makeatother

\def\paperID{***} 
\def\confName{CVPR}
\def\confYear{2026}

\title{Enhance-then-Balance Modality Collaboration for Robust Multimodal Sentiment Analysis}

\author{
    Kang He\textsuperscript{\rm 1,2}, 
    Yuzhe Ding\textsuperscript{\rm 1}, 
    Xinrong Wang\textsuperscript{\rm 1}, 
    Fei Li\textsuperscript{\rm 1}, 
    Chong Teng\textsuperscript{\rm 1}, 
    Donghong Ji\textsuperscript{\rm 1}\thanks{Corresponding author}\\
    \textsuperscript{\rm 1}Key Laboratory of Aerospace Information Security and Trusted Computing, Ministry of Education,\\
    School of Cyber Science and Engineering, Wuhan University. \\
    \textsuperscript{\rm 2}Shanghai Innovation Institute \\
{\tt\small \{hekang0225,yuzheding,dhji\}@whu.edu.cn}
}

\begin{document}
\maketitle
\begin{abstract}

Multimodal sentiment analysis (MSA) integrates heterogeneous text, audio, and visual signals to infer human emotions. While recent approaches leverage cross-modal complementarity, they often struggle to fully utilize weaker modalities. 
In practice, dominant modalities tend to overshadow non-verbal ones, inducing modality competition and limiting overall contributions. This imbalance degrades fusion performance and robustness under noisy or missing modalities.
To address this, we propose a novel model, \underline{E}nhance-then-\underline{B}alance \underline{M}odality \underline{C}ollaboration framework (EBMC). 
EBMC improves representation quality via semantic disentanglement and cross-modal enhancement, strengthening weaker modalities. 
To prevent dominant modalities from overwhelming others, an Energy-guided Modality Coordination mechanism achieves implicit gradient rebalancing via a differentiable equilibrium objective. 
Furthermore, Instance-aware Modality Trust Distillation estimates sample-level reliability to adaptively modulate fusion weights, ensuring robustness. 
Extensive experiments demonstrate that EBMC achieves state-of-the-art or competitive results and maintains strong performance under missing-modality settings.
Our codes are available at \href{https://github.com/kangverse/EBMC}{https://github.com/kangverse/EBMC}.
\end{abstract}    
\section{Introduction}
\label{sec:intro}
\begin{figure}[t]
    \centering \includegraphics[width=1.0\columnwidth]{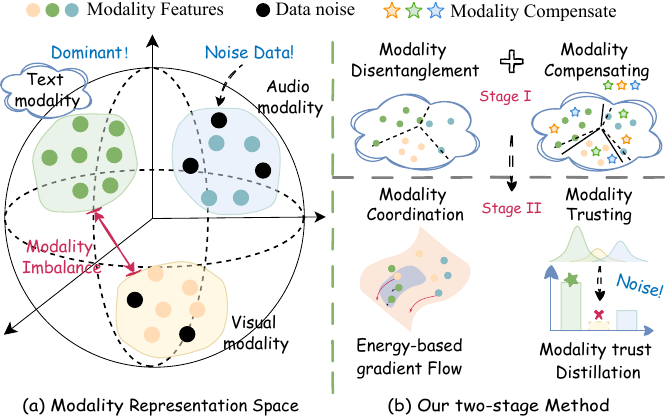}
    \caption{
    Illustration of modality imbalance: text tends to dominate learning while weaker modalities are easily suppressed and affected by noisy samples. 
    Our two-stage framework first performs modality disentanglement and compensation (Stage I), then applies an energy-based gradient flow for modality coordination and sample-aware reliability for trusting (Stage II), leading to balanced multimodal optimization and ensure robustness.
    }
    
    \label{fig:intro}
    \vspace{-5mm}
\end{figure}

Multimodal sentiment analysis (MSA) aims to capture the complex patterns of human emotional expression by integrating information from text, audio, and visual signals~\cite{poria2016convolutional,yu2019adapting,han2021improving,yu2023conki,fang2025catch}. 
Compared with unimodal approaches~\cite{wang-etal-2024-refining}, multimodal learning can exploit complementary cues across modalities, enabling more nuanced and robust emotion understanding in real-world scenarios~\cite{hazarika2020misa,lee2023multimodal}. 
With the widespread availability of multimodal data \cite{zadeh2016mosi,bagher-zadeh-etal-2018-multimodal,he-etal-2025-dalr}, MSA has shown promise in applications such as intelligent human–computer interaction \cite{zhang-etal-2024-escot,ding-etal-2025-zero} and mental health monitoring \cite{zhai-etal-2024-chinese}. 
However, due to the inherent heterogeneity across modalities and their unequal reliability~\cite{fernando2021missing,peng2022balanced,wu2022characterizing}, achieving robust sentiment analysis at a near-human cognitive level remains challenging.

Existing research has made substantial progress in multimodal representation learning and fusion.
Early~\cite{poria2016convolutional,yu2019adapting}, late~\cite{hazarika2020misa,yang2023confede,zhuang2024glomo}, and hybrid fusion strategies~\cite{zhang2025ecerc,li2025alignmamba,maniparambil2025harnessing} attempt to aggregate information at different stages of the pipeline.
Meanwhile, modality semantic disentanglement methods~\cite{li2023decoupled,wang2025dlf} isolate shared and modality-specific components, allowing models to better characterize the contribution of each modality to emotion expression.
These advances improve feature expressiveness and cross-modal interaction to some extent.
Nevertheless, most existing methods implicitly assume that \textit{all modalities contribute in a balanced and reliable manner}, and therefore offer \textbf{limited treatment of modality imbalance and robustness}.

Recent multimodal learning systems face a persistent challenge:
\textit{the expressive power and gradient contributions of different modalities are inherently imbalanced}.
Prior work~\cite{wang2020makes,fernando2021missing,fan2023pmr} shows that text consistently dominates prediction, while audio and visual signals remain under-exploited due to weaker or noisier emotional cues.
As illustrated in Fig.~\ref{fig:intro}(a), weaker modalities are more vulnerable to \textbf{noise data}, sparse cues, and missing signals, making them difficult to optimize and easily overshadowed during training.
This imbalance further induces \textit{modality competition}~\cite{fernando2021missing,peng2022balanced,fan2024detached,he2025pase}: dominant modalities accumulate larger gradients and reinforce their own representations, while weaker modalities receive insufficient updates.
Over time, this creates a “Matthew effect’’ where weak modalities become increasingly marginalized, \textit{especially under noisy or real-world conditions}.

To address these challenges, we propose a robust \underline{E}nhance-then-\underline{B}alance \underline{M}odality \underline{C}ollaboration framework (EBMC) with a two-stage design in Fig.~\ref{fig:intro}(b).
\textbf{Stage I} enhances unimodal representations: Modality Semantic Disentanglement (MSD) disentangles shared/specific semantics, and Cross-modal Complementary Enhancement (CCE) reinforces weak modalities through cross-modal complementary cues.
\textbf{Stage II} then balances modality contributions: Energy-guided Modality Coordination (EMC) module introduces an energy-based coordination mechanism that aligns modality optimization dynamics, and Instance-aware Modality Trust Distillation (IMTD) module further estimates modality reliability from probabilistic teacher signals to adaptively adjust fusion weights.

We conduct extensive experiments on three widely used benchmarks: CMU-MOSI, CMU-MOSEI, and IEMOCAP. 
EBMC achieves superior or highly competitive performance across all datasets. 
Given the close relationship between MSA and emotion recognition in conversations (ERC)~\cite{li2023revisiting}, we further transfer EBMC to ERC and observe consistent gains, demonstrating its generalizability. 
In addition, experiments under modality-missing and intra-modality missingness scenarios (~\cref{sec:analyses_discussion}) show that EBMC suffers substantially smaller performance degradation than existing baselines, confirming its robustness in more realistic, imperfect multimodal environments.
The main contributions are summarized as follows:
\begin{itemize}
    \item We introduce a weak-modality enhancement mechanism that disentangles shared/specific semantics and performs cross-modal compensation to strengthen weak modalities and improve emotion discrimination.
    \item To mitigate modality competition, we present the first EBM-inspired modality coordination mechanism that rebalances modalities via energy potentials and gradient-flow dynamics.
    \item We design an instance-aware modality trust distillation (IMTD) module that estimates sample-level modality reliability and adaptively modulates fusion weights, improving robustness under noisy or missing modalities.
    \item Extensive experiments on MOSI, MOSEI, and IEMOCAP demonstrate superior or comparable performance. We further validate EBMC on ERC tasks and under modality-missing scenarios, confirming its effectiveness. 

\end{itemize}

\section{Related works}
\label{sec:related_works}

\paragraph{Multimodal Sentiment Analysis.}
Multimodal Sentiment Analysis (MSA) integrates textual, acoustic, and visual cues to capture richer affective information. Existing approaches primarily fall into two directions: representation learning and multimodal fusion.
Representation learning methods focus on disentangling shared and modality-specific semantics, such as MISA \cite{hazarika2020misa}, as well as extensions using multimodal distillation \cite{li2023decoupled} and contrastive learning \cite{yang2023confede}.
Fusion approaches explore more adaptive strategies, including attention-based \cite{praveen2024recursive,shi2023multiemo}, graph-based \cite{xu2025towards,chen2025cr}, gating-based \cite{zhang2025ecerc}, and hierarchical models \cite{xu2025towards}. 
Recent work \cite{li2025alignmamba,fang2025emoe} further enhance modality alignment prior to fusion.
Although effective, these methods often overlook modality imbalance and struggle under noisy or missing modalities.

\vspace{-10pt}
\paragraph{Imbalanced Multimodal Learning.}
Modality imbalance occurs when dominant modalities overshadow weaker ones during joint training \cite{das2023revisiting,he2025pase}.
To alleviate this, prior works dynamically adjust learning rates or gradients across modalities based on loss signals or learning dynamics \cite{wang2020makes,wu2022characterizing,peng2022balanced}.
Other approaches redesign the objective to curb dominant modalities or boost weaker ones, including Fisher-information regularization \cite{huang2025adaptive}, decoupled fusion streams \cite{das2023revisiting}, detached unimodal encoders with contrastive objectives \cite{fan2024detached}, and prototype-based rebalancing \cite{fan2023pmr}.
While effective at the optimization level, these methods offer limited semantic enhancement for weaker modalities.

\vspace{-10pt}
\paragraph{Robust Multimodal Learning.}

Robust learning targets performance stability against modality noise, degradation, and missingness.
Uncertainty modeling methods, including hyperdimensional uncertainty quantification \cite{chen2025hyperdimensional} and aleatoric-guided fusion \cite{gao2024embracing}, improve confidence calibration and noise robustness, while prototype-based updating strategies further mitigate intra-modal variation \cite{li2025dpu}.
To handle missing modalities, recent works adopt self-distillation or cross-modal reconstruction \cite{li2024unified}, shared–specific representation learning \cite{lee2023multimodal}, or prompting-based alignment \cite{weng2025enhancing}.
Robustness is also enhanced through multi-scale attention fusion \cite{ma2023robust,tao2025multi} and text-centric adversarial prompting \cite{tsai2025enhance}.
In contrast, EBMC embeds robustness directly into the core learning process by adaptively suppressing noisy modalities and enhancing cross-modal complementarity for stable and reliable performance.
\section{Methodology}
\label{sec:method}

\subsection{Preliminaries}
\label{sec:preliminaries}

\begin{figure*}[t]
    \centering \includegraphics[width=2.0\columnwidth]{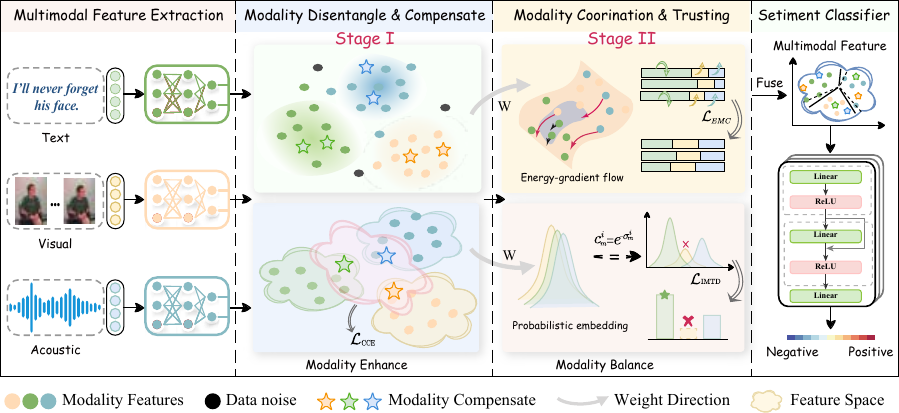}
    \caption{
    The overall architecture of our proposed model EBMC.}
    
    \label{fig:model}
    \vspace{-5mm}
\end{figure*}

In MSA task, the input comprises a set of modalities $M=\{l,v,a\}$ corresponding to text, visual, and audio signals.
Each modality is encoded as a feature sequence $X_m\in\mathbb{R}^{T_m\times d_m}$, where $T_m$ is the sequence length and $d_m$ the feature dimension.
Given multimodal inputs ${X_m}_{m\in M}$, the objective is to predict the sentiment category or intensity.
As illustrated in~\cref{fig:model}, our framework EBMC adopts a two-stage design: MSD (\cref{sec:MSD}) and CCE (\cref{sec:CCE}) first enhance weak modality representations while preserving unimodal predictive capability, and EMC (\cref{sec:emc}) together with IMTD (\cref{sec:IMTD}) then mitigates modality competition and strengthens robustness.

\subsection{Modality Semantic Disentangle}
\label{sec:MSD}
Directly fusing raw modality features often introduces semantic interference and allows dominant modalities to overshadow weaker ones, limiting the contribution of non-dominant modalities~\cite{wu2022characterizing,fan2023pmr,fan2024detached}.
To enable effective multimodal fusion while preserving unimodal predictive capability, we first perform semantic disentanglement for each modality.
Specifically, the representation of each modality $z_m$ is decomposed into a shared component $z_m^c$ and a modality-specific component $z_m^s$:
\vspace{-3pt}
\begin{equation}
z_{m}^{c}=D_{m}^{c}(z_{m})\mathrm{,~~~~}\quad z_{m}^{s}=D_{m}^{s}(z_{m})\mathrm{,}
\end{equation}
where $D_m^c$ and $D_m^s$ are lightweight MLP sub-networks that extract cross-modal shared semantics and modality-specific information (e.g., prosody or facial expression details). 
To ensure clear functional separation between shared and specific components, multiple constraints are introduced. First, invariant alignment ensures that shared components across modalities are close in the semantic space, formulated via an InfoNCE \cite{oord2018representation} contrastive loss:

\vspace{-4pt}
\begin{equation}
\mathcal{L}_{\mathrm{inv}}=-\sum_{i}\log\frac{\exp(\sin(z_{i}^{c},z_{\mathrm{agg}}^{c})/\tau)}{\sum_{k}\exp(\sin(z_{i}^{c},z_{k}^{c})/\tau)}.
\end{equation}
\vspace{-4pt}

Second, to avoid redundancy among modality-specific components, we minimize batch-wise cosine similarity, enforcing specific disentanglement:

\vspace{-5pt}
\begin{equation}
\mathcal{L}_{\mathrm{dis}}=\sum_{i\neq j}\sin(z_i^s,z_j^s).
\end{equation}
\vspace{-5pt}

In addition, to ensure each modality-specific component retains predictive capability, we employ a unimodal predictor $T_m$ and define the unimodal task loss as:
\vspace{-3pt}
\begin{equation}
\mathcal{L}_\mathrm{uni}=\frac{1}{|M|}\sum_{m\in M}\ell(T_m(z_m^s),y),
\end{equation}
where $\ell(\cdot,\cdot)$ is the task loss (e.g., cross-entropy or regression loss).
The overall loss for MSD is:
\vspace{-3pt}
\begin{equation}
\label{eq:msd}
\mathcal{L}_{\mathrm{MSD}}=\mathcal{L}_{\mathrm{inv}}+\lambda_1\mathcal{L}_{\mathrm{dis}}+\lambda_2\mathcal{L}_{\mathrm{uni}},
\end{equation}
where $\lambda_1$ and $\lambda_2$ are trade-off hyperparameters.
By disentangling modality representations in this way, MSD suppresses semantic interference and provides clean, controllable features for subsequent cross-modal collaboration.

\subsection{Cross-modal Complementary Enhancement}
\label{sec:CCE}

Although MSD preserves modality-specific information, the text modality still tends to dominate semantic discrimination, while visual and audio modalities often encode subtler emotional cues.
To better leverage these weaker modalities, we introduce the Cross-modal Complementary Enhancement (CCE) module, which enriches each modality by incorporating complementary shared and specific components from other modalities.

Concretely, for modality $m$, CCE takes its own components $(z_m^c,z_m^s)$ together with the shared and specific components from the remaining modalities $(z_{-m}^c,z_{-m}^s)$, and feeds them into a lightweight enhancement network $G_m$ to generate enhanced features:

\begin{equation}
\tilde{z}_m=G_m(z_m^c,z_{-m}^c,z_{-m}^s,\epsilon),
\end{equation}
where $\epsilon$ is an optional stochastic perturbation that increases feature diversity.
The training objective is designed to enforce semantic consistency and maintain downstream predictive correctness:
\vspace{-4pt}
\begin{equation}
\label{eq:cce}
\mathcal{L}_{\mathrm{CCE}}={\mathbb{E}\|\tilde{z}_m-z_m\|_2^2} +{\gamma\ell(f(\tilde{z}_m,z_{-m}),y)}.
\end{equation}
where $f(\cdot)$ denotes the downstream prediction head and $\gamma$ balances the two terms.
The reconstruction term preserves the original semantic structure of $z_m$, while the task loss ensures that the enhanced feature remains discriminative for the target task.
In this way, CCE effectively amplifies the contribution of weaker modalities and provides more informative representations for subsequent EMC module.

\subsection{Energy-guided Modality Coordination}
\label{sec:emc}
To further address modality competition, we reinterpret multimodal coordination through the lens of Energy-based Models (EBMs). 
Rather than explicitly reweighting losses or manually correcting gradients, EMC constructs a structured multimodal energy landscape and performs coordination through both energy-gap minimization and energy-gradient dynamics. 
This provides a fully differentiable mechanism for suppressing dominant modalities and amplifying weaker ones during joint optimization.

\textbf{Energy Function.}
For each modality $m$, we define a modality-specific energy potential that integrates semantic activation, task difficulty, and predictive reliability:
\begin{equation}
E(m)= 
\alpha \| z_m \|_2^2 
+ \beta\, \ell_m
+ \gamma\, u_m ,
\label{eq:emc_energy}
\end{equation}
\begin{equation}
u_m = \mathbb{E}_i \!\left[ H\!\left(p_{T_m}^i(y)\right) \right], 
H(p) = - \sum_{y} p(y)\log p(y).
\end{equation}
where $\|z_m\|_2^2$ captures representation magnitude, $\ell_m$ is the modality-specific task loss, and $u_m$ denotes predictive uncertainty.
This formulation yields a multi-factor energy landscape in which weaker modalities typically exhibit higher energy due to noisier signals or less discriminative features.
Refer to Appendix~\ref{appendix:emc_theory} for details.

\textbf{Energy-gap Minimization.}
Dominant modalities often acquire disproportionately low energy, enabling them to overwhelm weaker ones. 
To prevent this imbalance, EMC enforces a global energy-equilibrium objective:
\vspace{-3pt}
\begin{equation}
\mathcal{L}_{gap}
= \sum_{i,j} (E(m_i)-E(m_j))^2.
\label{eq:emc_gap}
\end{equation}
Minimizing~\cref{eq:emc_gap} encourages all modalities to converge toward a balanced regime, ensuring that under-utilized modalities receive increased optimization focus.

\textbf{Energy-gradient Flow.}
Beyond matching energy magnitudes, EMC further regulates optimization dynamics by explicitly adopting an energy-descent update:
\begin{equation}
\Delta z_m
= - \lambda \frac{\partial E(m)}{\partial z_m}.
\label{eq:emc_flow}
\end{equation}
This implicitly forms a negative-feedback mechanism:  
modalities with excessively low energy (over-confident or overly dominant) receive suppressive gradients, whereas high-energy modalities naturally obtain larger corrective gradients.
Such dynamics approximate gradient flows in EBMs and provide an adaptive, self-correcting training trajectory. 
Combining the above components, the complete EMC objective becomes:
\begin{equation}
\mathcal{L}_{EMC}
= \mathcal{L}_{gap}
+ \delta \sum_{m}\left\|\frac{\partial E(m)}{\partial z_m}\right\|^2,
\label{eq:emc_final}
\end{equation}
where the second term penalizes unstable or overly sharp energy gradients, promoting smoother energy flows and stable multimodal convergence.

\subsection{Instance-aware Modality Trust Distillation}
\label{sec:IMTD}

MSD and CCE alleviate modality imbalance at the representation level; 
however, traditional fusion schemes remain vulnerable to noise and missing modalities at the instance level~\cite{lian2023gcnet,li2024correlation,zhang2024towards}.
To further enhance robustness, we introduce Instance-aware Modality Trust Distillation (IMTD), which uses probabilistic embeddings to estimate sample-specific modality confidence and adaptively weight modality contributions during distillation.

Specifically, for modality $m$, the teacher model $T_m$ from MSD (\cref{sec:MSD}) produces a predictive probability distribution $p_{T_m}(y)$ for each sample.
We denote its mean and variance as $\mu_m^i$ and $\sigma_m^{i}$, where $\sigma_m^{i}$ represents the predictive uncertainty of modality $m$ for sample $i$.
We convert the variance into a confidence score:
\vspace{-2pt}
\begin{equation}
c_m^i = \exp(-\sigma_m^i),
\end{equation}
so lower variance implies higher certainty and thus higher confidence.
To avoid instability from excessively large variances, we introduce a soft normalization factor:
\begin{equation}
\rho_m^i = \frac{1}{\log(1 + \|\sigma_m^{2^i}\|_1)},
\end{equation}
which suppresses the influence of unstable modalities and smooths the confidence distribution.
The final adaptive distillation weight combines teacher confidence and normalized reliability:
\vspace{-2pt}
\begin{equation}
\alpha_m^i = \frac{c_m^i \rho_m^i}{\sum_m c_m^i \rho_m^i}.
\end{equation}

During distillation, the student’s fused prediction is aligned with the teacher outputs via a confidence-weighted KL divergence:
\vspace{-3pt}
\begin{equation}
\label{eq:imtd}
\mathcal{L}_{\mathrm{IMTD}} = \sum_{m,i} \alpha_m^i 
\operatorname{KL}\Big(\sigma(z_{\mathrm{fusion}}^i / \tau) \parallel \sigma(z_{T_m}^i / \tau)\Big),
\end{equation}
where $\sigma(\cdot)$ denotes the softmax function and $\tau$ is a temperature coefficient.
IMTD a\textit{ssigns higher weights to reliable modalities} and \textit{down-weights noisy ones} at the sample level, providing a stable way to handle modality missingness and noise while enhancing robustness; further analyses will be discussed in ~\cref{sec:analyses_discussion}.

\subsection{Overall Training Objective}
With the losses defined in \cref{eq:msd} – \cref{eq:imtd}, the overall EBMC training objective is:
\vspace{-3pt}
\begin{equation}
\mathcal{L}
=\mathcal{L}_{\mathrm{task}}
+\zeta \mathcal{L}_{\mathrm{MSD}}
+\beta \mathcal{L}_{\mathrm{CCE}}
+\gamma \mathcal{L}_{\mathrm{EMC}}
+\eta \mathcal{L}_{\mathrm{IMTD}}.
\label{eq:total_loss}
\end{equation}

\begin{equation}
    \mathcal{L}_\mathrm{task} = \frac{1}{N} \sum_{i=1}^{N} - y_i \log(\hat{y}_i)
\end{equation}
This objective jointly drives disentanglement, complementary enhancement, balanced coordination, and reliability-aware fusion, with $\zeta, \beta,\gamma,\eta$ weighting each component.
\section{Experiments}
\label{sec:experiments}

\vspace{-0.3mm}
\paragraph{Datasets.}
To evaluate the generalizability of our framework for both MSA and ERC tasks, we conduct experiments on three widely used benchmark datasets: CMU-MOSI \cite{zadeh2016mosi}, CMU-MOSEI \cite{bagher-zadeh-etal-2018-multimodal}, and IEMOCAP \cite{busso2008iemocap}.
CMU-MOSI contains 2,199 opinion video clips annotated with binary sentiment labels. 
CMU-MOSEI is a larger and more diverse dataset with 22,856 YouTube video segments.
Both datasets are labeled with sentiment intensity scores from $-3$ (strongly negative) to $+3$ (strongly positive), enabling fine-grained regression-based evaluation. 
IEMOCAP, a standard benchmark for emotion recognition, consists of 4,453 video clips annotated with categorical emotions across approximately 12 hours of audiovisual recordings.

\vspace{-0.3cm}
\paragraph{Evaluation Metrics.}
We report the average results over five runs with different random seeds. 
For MOSI and MOSEI, we evaluate EBMC performance using metrics including 2-class Accuracy (Acc-2), 7-class Accuracy (Acc-7), F1 score (F1), Correlation (Corr), and Mean Absolute Error (MAE). 
Following prior work conventions \cite{yu2023conki,zhuang2024glomo}, Acc-2 and F1 are reported in two forms (denoted as `-/-'): one with zero as negative/non-negative, and the other excluding zero as negative/positive. 
For IEMOCAP, as recommended by \cite{wang2019words,li2024correlation}, we evaluate the F1 score for four emotion categories (\textit{happy}, \textit{sad}, \textit{angry}, and \textit{neutral}).

\vspace{-0.3cm}
\paragraph{Implementation Details.}
To ensure fair comparisons with previous studies, we use the same feature extraction settings. 
For language modality, we use \texttt{BERT$_\textrm{base}$} model~\cite{devlin2019bert} to obtain a 768-dimensional hidden state as the word features, and additionally incorporate 300-dimensional GloVe features~\cite{pennington-etal-2014-glove} for consistency with prior ERC work on IEMOCAP.
For the visual modality, we extract 35 facial action unit features using Facet, 
and for the acoustic modality, we use COVAREP~\cite{degottex2014covarep} to obtain 74-dimensional low-level acoustic descriptors. 
Hyperparameters $\lambda_1,\lambda_2$, $\beta,\gamma,\eta$ are all set to 0.1, and $\zeta$ is set to 0.5 based on validation performance. 
All experiments are conducted on a RTX 4090 GPU (48GB) with a batch size of 64.
Each stage of EBMC is trained for 100 epochs. 
Additional implementation details are provided in Appendix~\ref{appendix:more_implementation_details}.

\subsection{Comparison with State-of-the-art Methods}
\label{sec:main_results}
We conduct a comprehensive comparison between our proposed EBMC and several representative SOTA models, including MuLT \cite{tsai2019multimodal}, SelfMM\cite{yu2021learning}, ConKI \cite{yu2023conki}, ConFEDE \cite{yang2023confede}, CLGSI~\cite{yang2024clgsi}, MFON~\cite{zhang2025modal}, EUAR \cite{gao2024enhanced}, EMOE~\cite{fang2025emoe},
GLoMo \cite{zhuang2024glomo}, DEVA~\cite{wu2025enriching}, Semi-IIN~\cite{lin2025semi}. 
In addition, to evaluate the robustness of EBMC, we perform extensive experiments under modality-missing scenarios~\cite{lian2023gcnet,li2024correlation,li2024unified}. Further details can be found in Appendix~\ref{appendix:more_baselines}.

\begin{table*}[t]

\centering
\caption{Comparison with SOTA methods on the CMU-MOSI and CMU-MOSEI datasets. Bold indicates the best result, and underlining denotes the second-best. 
Note: $\heartsuit$ and $\diamondsuit$ represents results obtained from \cite{zhuang2024glomo} and \cite{yu2023conki}, respectively. $\ast$ indicates reproduced results from the public code. Other results taken directly from the original paper. See details in Sec.~\ref{sec:main_results}.}
\begin{adjustbox}{width=2.0\columnwidth}
\begin{tabular}{l|cccccccccccc}
\toprule
\multirow{2}{*}{Method} & \multicolumn{5}{c}{CMU-MOSI} & \multicolumn{5}{c}{CMU-MOSEI} \\
\cmidrule(lr){2-6} \cmidrule(lr){7-11}
 &  Acc-2($\uparrow$) & F1($\uparrow$) & Acc-7($\uparrow$) & Corr($\uparrow$) & MAE($\downarrow$) & Acc-2($\uparrow$) & F1($\uparrow$) & Acc-7($\uparrow$) & Corr($\uparrow$) & MAE($\downarrow$)\\
\midrule

MuLT$^\heartsuit$~\cite{tsai2019multimodal} & 79.51/80.47 & 79.46/80.49 & 36.74 & 0.667 & 0.892 &	81.10/83.63 & 81.05/83.46 & 52.34 & 0.605 & 0.671 \\

SelfMM$^\diamondsuit$~\cite{yu2021learning}& 82.54/84.77 & 84.42/85.95 & 45.79 & 0.795 & 0.712 & 82.68/84.96 & 82.95/84.93 & 53.46 & 0.767 & 0.529 \\

ConKI$^\diamondsuit$~\cite{yu2023conki} & 84.37/86.13 & 84.33/86.13 & 48.43 & 0.816 & \underline{0.681} & 82.73/86.25 & 83.08/86.15 & 54.25 & 0.782 & 0.529 \\

ConFEDE~\cite{yang2023confede} & 84.17/85.52 & 84.13/85.52 & 42.27 & 0.742 & 0.784 & 81.65/85.82 & 82.17/85.83 & 54.86 & 0.780 & 0.522 \\

CLGSI~\cite{yang2024clgsi} & 83.97/86.43 & 83.63/86.25 & 47.96 & 0.790 & 0.703 & 84.01/86.32 & 84.21/86.18 & 54.56 & 0.763 & 0.532 \\

MFON~\cite{zhang2025modal} & 84.84/86.89 & 84.75/86.86 & 44.90 & 0.797 & 0.725 & 82.70/86.32 & 83.13/86.29 & 53.72 & 0.780 & 0.528 \\

EUAR$^\ast$~\cite{gao2024enhanced} & 83.24/84.58 & 84.22/84.58 & 48.18 & 0.789 & 0.711 & 82.20/85.64 & 82.02/85.46 & 51.07 & 0.785 & 0.596 \\

EMOE$^\ast$~\cite{fang2025emoe} & 83.73/85.58 & 83.68 /85.59 & 38.16 & 0.798 & 0.736 & 81.90/85.66 & 82.51/85.69  & 52.89 & 0.708 & 0.612 \\

GLoMo$^\heartsuit$~\cite{zhuang2024glomo} & 84.10/86.70 & 83.90/86.60 & 48.30 & 0.782 & 0.718 & 83.70/86.50 & 84.00/86.40 & 55.00 & 0.771 & 0.539 \\

DEVA~\cite{wu2025enriching} & 84.40/86.29 & 84.48/86.30 & 46.32 & 0.787 & 0.730 & 83.26/86.13 & 82.93/86.21 & 52.26& 0.769 & 0.541 \\
Semi-IIN~\cite{lin2025semi} &85.28/87.04 & 85.19/87.00 & 46.50 & \underline{0.822} & \textbf{0.679} & 84.98/\underline{87.70} & 85.27/\underline{87.65} & \underline{55.89}  & \underline{0.804} & \textbf{0.497} \\

EBMC(\textbf{ours}) & \textbf{86.26/87.84} & \textbf{86.20/87.79} & \textbf{50.34} & \textbf{0.833} & 0.694 & \textbf{86.04/88.10} & \textbf{86.23/88.07} & \textbf{57.32} & \textbf{0.824} & 0.505 \\

\bottomrule
\end{tabular}
\end{adjustbox}

\label{tab:main_results_mosi_mosei}
\vspace{-4mm}

\end{table*}

\vspace{-0.3cm}
\paragraph{Results on the MSA dataset.}
\cref{tab:main_results_mosi_mosei} demonstrates that EBMC consistently outperforms existing SOTA methods across most evaluation metrics on the CMU-MOSI and CMU-MOSEI datasets. 
Notably, EBMC demonstrates marked improvements in Acc-7, outperforming Semi-IIN~\cite{lin2025semi} by 1.89\% on CMU-MOSI (50.34\% vs. 46.50\%) and by 1.1\% on CMU-MOSEI (57.32\% vs. 55.89\%). 
This indicates that the complementary enhancement of weaker modalities allows the model to \textit{better capture discriminative information} in samples with subtle emotional expressions. 
Furthermore, improvements are observed across other metrics; 
compared with GLoMo~\cite{zhuang2024glomo}, EBMC achieves a 1.19\% higher F1 score on CMU-MOSI (87.79\% vs. 86.60\%) and a 1.67\% higher F1 score on CMU-MOSEI (88.07\% vs. 86.40\%). 
Overall, EBMC effectively disentangles shared and modality-specific information and enhances weaker modalities through the CCE module. 
In addition, we also employ the energy-based modality coordination mechanism to dynamically balance contributions across modalities, leading to consistently improved performance.




\begin{table}[t]
\centering
\caption{Comparison of F1-scores across different methods on the IEMOCAP dataset. Please see~\cref{sec:main_results} for details.}
\begin{adjustbox}{width=0.85\columnwidth}
\begin{tabular}{l|ccccc}
\toprule
\textbf{Methods} & \textbf{Happy} & \textbf{Sad} & \textbf{Angry} & \textbf{Neutral} & \textbf{Avg.} \\
\midrule
Self-MM~\cite{yu2021learning} & 90.8 & 86.7 & 88.4 & 72.7 & 84.65 \\
MCTN~\cite{pham2019found} & 83.1 & 82.8 & 84.6 & 67.7 & 79.55 \\
TransM~\cite{wang2020transmodality} & 85.5 & 84.0 & 86.1 & 67.1 & 80.68 \\
SMIL~\cite{ma2021smil} & 86.8 & 85.2 & 84.9 & 68.9 & 81.45 \\
GCNet~\cite{lian2023gcnet} & 87.7 & 86.9 & 85.2 & 71.1 & 82.73 \\
UMDF~\cite{li2024unified} & 87.9 & 86.5 & 85.8 & 70.5 & 82.68 \\
CorrKD~\cite{li2024correlation} & 87.5 & 85.9 & 86.1 & 71.5 & 82.75 \\
DMD~\cite{li2023decoupled} & 91.1 & 88.4 & 88.6 & 72.2 & 85.08 \\

EBMC (\textbf{Ours}) & \textbf{92.0} & \textbf{89.5} & \textbf{90.3} & \textbf{73.6} & \textbf{86.35} \\
\bottomrule
\end{tabular}
\end{adjustbox}
\label{tab:experiments_iemocap}
\vspace{-3mm}
\end{table}
\vspace{-0.3cm}
\paragraph{Results on the ERC dataset.}
To further evaluate the generalization ability of EBMC, we assess its performance on the ERC task. 
As shown in \cref{tab:experiments_iemocap}, EBMC demonstrates strong performance on the IEMOCAP dataset across all four emotion categories. 
The strongest baseline DMD ~\cite{li2023decoupled}, records an average F1-score of 85.08\%, whereas EBMC achieves 86.35\%, reflecting a 1.27\% improvement.
Notably, ERC involves finer-grained emotion categories, where human emotional expression can emphasize different cues, and certain signals may be more salient in audio or visual modalities. 
Moreover,~\cref{tab:experiments_iemocap} also shows that EBMC achieves particularly pronounced gains in \textit{Happy} and \textit{Angry} category, indicating enhanced discriminative capability in capturing emotional intensity and semantic details. 
Overall, EBMC can excel high-level performance on fine-grained, complex ERC scenarios, further validating the effectiveness of its complementary enhancement and energy-based modality coordination mechanisms.

\begin{figure*}[t]
    \centering \includegraphics[width=1.95 \columnwidth]{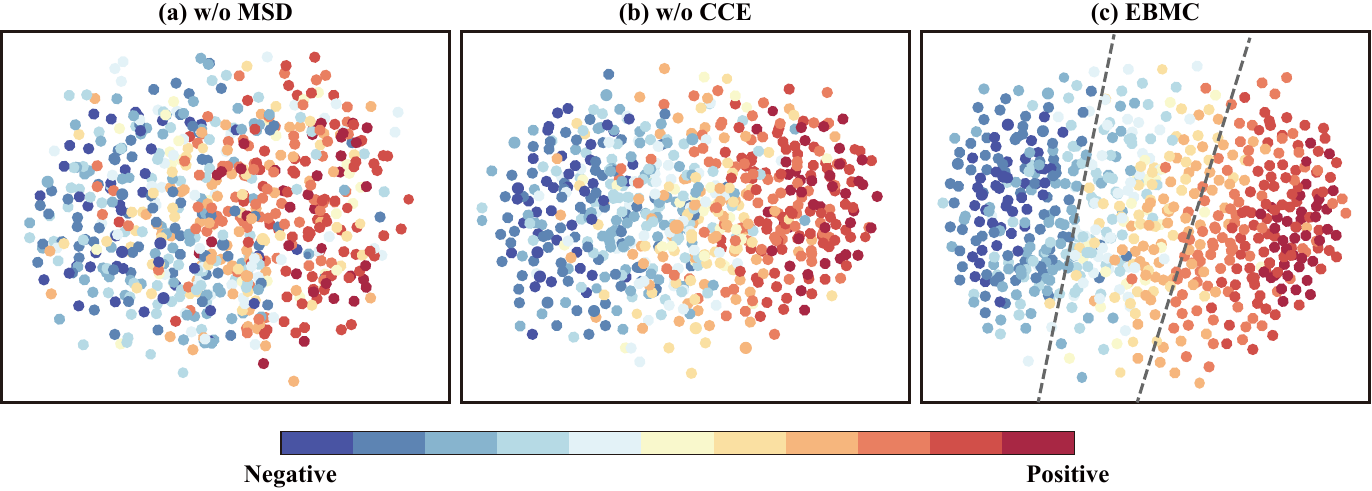}
    \caption{
    T-SNE visualization of features distrubution on CMU-MOSI. The closer color is to red, the more positive sentiment. EBMC demonstrates superior performance in MSA. 
    Please refer to \cref{sec:analyses_discussion}.}
    
    \label{fig:ablation_tsne}
    \vspace{-3mm}
\end{figure*}

\subsection{Robust Analysis on Incomplete Conditions }
\label{sec:robust_analysis}
To comprehensively evaluate the performance of EBMC under various incomplete modality conditions, we compared it with other methods \cite{pham2019found,lian2023gcnet,gao2024enhanced} on the CMU-MOSEI dataset. 
As shown in~\cref{tab:modality_missing_mosei_without_atv}, EBMC achieves the highest F1 scores under almost all missing modality conditions. 
We observe the following: 
(i) Competitor methods \textit{\textbf{perform significantly worse in single-modal conditions}}, especially when only visual or audio modalities are available. EBMC effectively mitigates this issue. 
In the initial stage, MSD and CCE modules are able to learn useful information even with missing modalities. 
(ii) Modality missingness leads to a decline in performance across all modalities, which further highlights the \textbf{\textit{heterogeneity between modalities}}. 
As the number of modalities increases, integrating complementary information from different modalities can effectively enhance the emotional semantics in the joint representation.
These results demonstrate the superiority of EBMC in multi-modal missingness scenarios, enabling the model to fully leverage the potential of remaining modalities and compensate for the impact of missing information.

\begin{table}[t]
\centering
\caption{Comparison of performance under six incomplete modality testing conditions on CMU-MOSEI datasets. Refer to Sec.~\ref{sec:robust_analysis}.
}
\begin{adjustbox}{width=1.0\columnwidth}
\begin{tabular}{l|cccccccc}
\toprule
\multirow{2}{*}[-1.4mm]{\textbf{Models}} & \multicolumn{7}{c}{\textbf{Testing Conditions}} \\
\cmidrule{2-8} 
& \textbf{\{A\}} & \textbf{\{T\}} & \textbf{\{V\}} & \textbf{\{A,V\}} & \textbf{\{A,T\}} & \textbf{\{T,V\}} & \textbf{Avg.} \\
\midrule

Self-MM~\cite{yu2021learning}& 43.57 & 71.53 & 37.61 & 49.52 & 75.91 & 74.62 & 58.79 \\
CubeMLP~\cite{sun2022cubemlp} & 39.54 & 67.52 & 32.58 & 48.54 & 71.69 & 70.06 & 54.99 \\
 
MCTN~\cite{pham2019found} & 62.72 & 75.50 & 59.46 & 64.84 & 76.64 & 77.13 & 69.38 \\
TransM~\cite{wang2020transmodality} & 63.68 & 77.98 & 58.67 & 62.24 & 80.46 & 78.61 & 70.27 \\
GCNet~\cite{lian2023gcnet}  & \underline{66.54} & 80.52 & 61.83 & 69.21 & 81.96 & 81.15 & 73.54 \\
DMD~\cite{li2024correlation} & 46.18 & 70.26 & 39.84 & 52.70 & 74.78 & 72.45 & 59.37 \\
CorrKD~\cite{li2024correlation} & 66.09 & 80.76 & 62.30 & \underline{71.92} & 81.74 & 81.28 & 74.02 \\
EUAR~\cite{gao2024enhanced} & 60.70 & \textbf{85.20} & \underline{65.30} & 65.30 & \underline{85.10} & \underline{86.00} & \underline{74.50} \\
EBMC(\textbf{Ours}) & \textbf{71.67} & \underline{84.79} & \textbf{70.05} & \textbf{73.34} & \textbf{86.76} & \textbf{86.90} & \textbf{78.92} \\

\bottomrule

\end{tabular}

\label{tab:modality_missing_mosei_without_atv}
\end{adjustbox}
\vspace{-4mm}

\end{table}

\subsection{Ablation Studies}
\label{sec:ablation_studies}

To assess the contribution of each key component in EBMC, we conduct ablation studies on the CMU-MOSI and CMU-MOSEI datasets. 
\textbf{Firstly}, as shown in~\cref{tab:ablation_module}, removing the MSD module significantly impairs the model's ability to disentangle shared and modality-specific information, leading to a decrease in F1 by 1.63\% on CMU-MOSI and 1.76\% on CMU-MOSEI. 
\textbf{Similarly}, omitting the CCE module weakens the enhancement of weaker modalities, resulting in F1 reductions of 0.89\% and 1.18\%, respectively. 
The exclusion of the EMC module severely disrupts modality coordination, causing the largest F1 drop of 2.43\% and 2.87\%. 
\textbf{Finally}, removing the IMTD module, which ensures the adaptive weighting of modalities based on sample-level reliability, leads to a reduction of 1.02\% and 0.98\% in F1 on the two datasets. 
IMTD is crucial for modality missingness, as further discussed in~\cref{sec:analyses_discussion} (Q3).
These findings highlight the critical role of each component in EBMC, emphasizing their individual contributions to improving discriminative multimodal feature learning and overall robustness.

\begin{table}[t!]
\centering
\caption{Ablation studies for each module in our proposed EBMC on CMU-MOSI and CMU-MOSEI datasets. Refer to Sec.~\ref{sec:ablation_studies}.}
\begin{adjustbox}{width=1.00\columnwidth}
\begin{tabular}{l|cc|cc}
\toprule
\multirow{2}{*}{Model} & \multicolumn{2}{c}{\textbf{CMU-MOSI}} & \multicolumn{2}{c}{\textbf{CMU-MOSEI}} \\
\cmidrule(lr){2-3} 
\cmidrule(lr){4-5}  
 & Acc-2($\uparrow$) & F1($\uparrow$)  & Acc-2($\uparrow$) & F1($\uparrow$) \\
\midrule

\textbf{EBMC} & \textbf{86.26/87.84} & \textbf{86.20/87.79} & \textbf{86.04/88.10} & \textbf{86.23/88.07} \\
\quad w/o MSD & 84.37/86.23 & 84.30/86.16 & 84.24/86.32 & 84.16/86.31 \\
\quad w/o CCE & 84.92/86.98 & 84.87/86.90 & 84.55/86.96 & 84.52/86.89 \\
\quad w/o EMC & 83.54/85.41 & 83.56/85.32 & 83.51/85.12 & 83.47/85.20 \\
\quad w/o IMTD & 85.35/87.01 & 85.24/86.77 & 85.26/87.15 & 84.96/87.09 \\
\midrule

\end{tabular}
\end{adjustbox}
\label{tab:ablation_module}
\vspace{-3mm}
\end{table}

\begin{table*}[htbp]
\centering
\caption{Robustness comparison of overall performance on the CMU-MOSI and CMU-MOSEI datasets under intra-modality missingness, simulated by randomly dropping frame-level features at varying rates ($p\in\{0, 0.1, \ldots, 0.9\}$). Refer to Sec.~\ref{sec:analyses_discussion} for details.}

\begin{adjustbox}{width=1.98\columnwidth}
\begin{tabular}{l|cccccccccccc}
\toprule
\multirow{2}{*}{Methods} & \multicolumn{5}{c}{\textbf{CMU-MOSI}} & \multicolumn{5}{c}{\textbf{CMU-MOSEI}} \\
\cmidrule(lr){2-6} \cmidrule(lr){7-11}
& Acc-2($\uparrow$) & F1($\uparrow$) & Acc-7($\uparrow$) & Corr($\uparrow$) & MAE($\downarrow$) & Acc-2($\uparrow$) & F1($\uparrow$) & Acc-7($\uparrow$) & Corr($\uparrow$) & MAE($\downarrow$)\\
\midrule
MISA~\cite{hazarika2020misa} & 70.33/71.49 & 70.00/71.28 & 29.85 & 0.524 & 1.085 & 75.82/71.27 & 68.73/63.85 & 40.84 & 0.503 & 0.780 \\
Self-MM~\cite{yu2021learning} & 69.26/70.51 & 67.54/66.60 & 29.55 & 0.512 & 1.070 & 77.42/73.89 & 72.31/68.92 & 44.70 & 0.498 & 0.695 \\
MMIM~\cite{han2021improving} & 67.06/69.14 & 64.04/66.65 & 31.30 & 0.507 & 1.077 & 75.89/73.32 & 70.32/68.72 & 40.75 & 0.489 & 0.739 \\
CENET~\cite{wang2022cross} & 67.73/71.46 & 64.85/68.41 & 30.38 & 0.504 & 1.080 & 77.34/74.67 & 74.08/70.68 & 47.18 & 0.535 & 0.685 \\
TETFN~\cite{wang2023tetfn} & 67.68/69.76 & 63.29/65.69 & 30.30 & 0.507 & 1.087 & 67.68/69.76 & 63.29/65.69 & 30.30 & 0.508 & 1.087 \\
TFR-Net~\cite{yuan2021transformer} & 66.35/68.15 & 60.06/61.73 & 29.54 & 0.459 & 1.200 & 77.23/73.62 & 71.99/68.80 & 46.83 & 0.489 & 0.697 \\
ALMT~\cite{zhang2023learning} & 68.39/70.40 & 71.80/72.57 & 30.30 & 0.498 & 1.083 & 77.54/76.64 & 78.03/77.14 & 40.92 & 0.481 & \textbf{0.674} \\
LNLN~\cite{zhang2024towards} & 70.94/72.55 & 71.25/72.73 & 34.26 & 0.527 & 1.046 & 78.19/76.30 & 79.95/77.77 & 45.42 & 0.530 & 0.692 \\
EBMC(\textbf{Ours}) & \textbf{72.30/73.48} & \textbf{72.19/74.25} & \textbf{34.65} & \textbf{0.545} & \textbf{0.994} & \textbf{78.56/76.93} & \textbf{80.17/78.32} &\textbf{45.69} & \textbf{0.588} & 0.680 \\
\bottomrule
\end{tabular}
\end{adjustbox}
\label{tab:robustness_comparison}
\vspace{-3mm}
\end{table*}

\subsection{Analyses and Discussions }
\label{sec:analyses_discussion}
To better assess the effectiveness of EBMC, we conduct in-depth analyses to answer the following key questions: 

\vspace{1mm}
\textbf{Q1: How does weak modality enhancement improve emotion classification performance?}
To evaluate the effectiveness of MSD and CCE, we conduct ablation studies and use t-SNE~\cite{maaten2008visualizing} for 2D visualization of high-dimensional features. \cref{fig:ablation_tsne}a (w/o MSD) shows a mixed feature distribution with unclear emotional boundaries, indicating that without semantic alignment, the model struggles to fuse information, resulting in scattered features and low emotional category distinction. 
Introducing MSD (\cref{fig:ablation_tsne}b) improves feature distribution, making emotional boundaries clearer, though scattered points remain in transition areas due to weak modality signals in ambiguous samples. 
MSD aligns shared semantic components but doesn’t fully compensate for weak modality signals. With CCE (\cref{fig:ablation_tsne}c), scattered points cluster into corresponding emotional categories, enhancing weak modality representation and improving emotion classification accuracy. 
These results demonstrate that weak modality enhancement addresses modality heterogeneity, enhances weak modality representation, and improves classification accuracy.

\begin{figure}[t]
    \centering \includegraphics[width=1.0 \columnwidth]{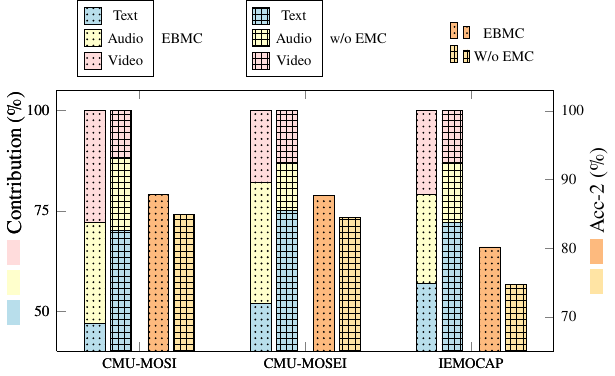}
    \caption{
    The impact of EMC on modality contributions and overall performance. The dual y-axes show the average contribution of each modality (left) and the corresponding Acc-2 score (right). Please refer to Sec.~\ref{sec:analyses_discussion}. Best viewed in color.}
    
    \label{fig:ablation_EMC}
    \vspace{-6mm}
\end{figure}

\vspace{1mm}
\textbf{Q2: How does energy-driven modality collaboration alleviate the modality competition issue?}
EMC addresses modality competition by dynamically balancing the contributions of each modality during training.
As shown in \cref{fig:ablation_EMC}, without EMC, the textual modality dominates with over 50\% contribution on all three datasets, suppressing audio and video signals and leading to imbalanced multimodal fusion.
With EMC, the contribution distribution across modalities improves significantly: text dominance is reduced while audio and video contributions increase, resulting in a more balanced utilization of all modality information.
This rebalancing directly translates to performance gains, as reflected by consistently higher Acc-2 scores across MOSI, MOSEI, and IEMOCAP.
These results demonstrate that EMC effectively reduces modality competition, ensuring weaker modalities are not overshadowed, and yielding a more comprehensive representation that enhances performance in both MSA and ERC tasks.

\vspace{1mm}
\textbf{Q3: What are the advantages of instance-aware modality trust distillation?}
We follow prior works~\cite{yuan2021transformer,zhang2024towards} by randomly dropping frame-level features at varying rates ($p\in\{0, 0.1, \ldots, 0.9\}$) to simulate intra-modality missingness, and average results across all rates to measure robustness.
As shown in \cref{tab:robustness_comparison}, EBMC consistently outperforms all baselines on both datasets.
On CMU-MOSI, it surpasses LNLN~\cite{zhang2024towards} by 1.52\% in F1 and 0.39\% in Acc-7, while reducing MAE from 1.046 to 0.994; similar gains hold on CMU-MOSEI, with Corr improving by 0.058.
This robustness stems from the IMTD mechanism, which dynamically estimates per-sample modality reliability at inference time.
When a modality is heavily corrupted, IMTD suppresses its contribution and up-weights more reliable modalities, enabling fine-grained rebalancing that fixed-weight or modality-level fusion strategies~\cite{yuan2021transformer,zhang2024towards} cannot achieve.
As a result, EBMC retains discriminative cross-modal information under high noise and yields better-calibrated sentiment intensity predictions.

\begin{figure}[t]
\centering 
\includegraphics[width=1.0 \columnwidth]{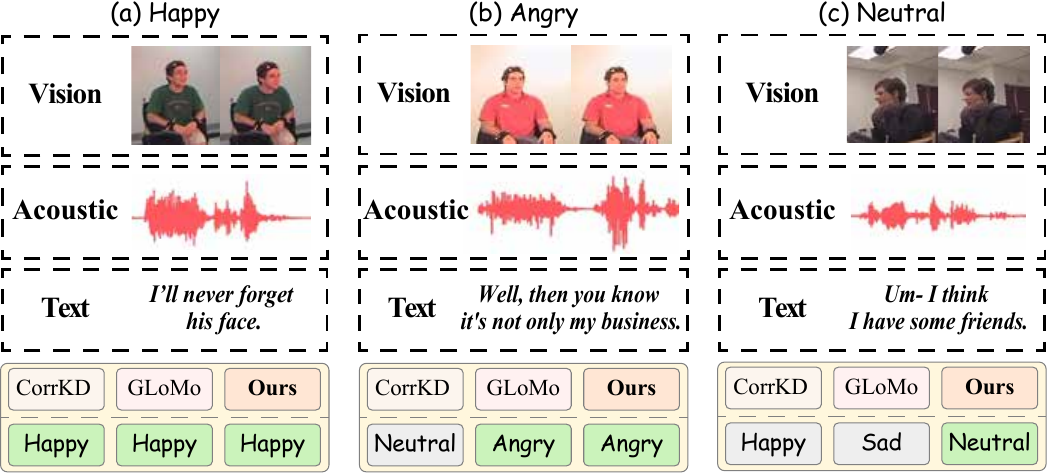}
\caption{
Comparison of EBMC and baseline predictions on the test sets of IEMOCAP. Refer to ~\cref{sec:case_study}.
}
\label{fig:case_study}
\vspace{-3mm}
\end{figure}

\subsection{Case Study}
\label{sec:case_study}

We compare EBMC with strong baselines on IEMOCAP samples.
For clear emotional expressions (\cref{fig:case_study}a), all models correctly recognize the sentiment.
When emotional intensity is low—for example, neutral utterances with flat prosody or weakly negative expressions with minimal facial cues—baseline models tend to over-rely on textual polarity.
As a result, they often mispredict the emotion, as shown in~\cref{fig:case_study}c where the true Neutral label is incorrectly classified as Happy or Sad.
In contrast, EBMC leverages enhanced audio–visual signals through CCE and mitigates textual dominance via EMC, enabling the model to capture weak prosody, subtle facial movements, and other non-verbal markers that better reflect the true emotional state.
Overall, these cases demonstrate EBMC’s superiority in discerning subtle emotions, resolving cross-modal imbalance, and achieving robust multimodal inference.

\section{Conclusion}
\label{sec:conclusion}

In this work, we present a simple yet effective Enhance-then-Balance Modality Collaboration (EBMC) framework, suitable for both full-modality and modality-missing scenarios. 
EBMC adopts a two-stage design: the first stage extracts and aligns semantic information from each modality, preserving unimodal peak prediction capability; 
the second stage introduces adaptive fusion using EMC and IMTD to optimize the contributions of each modality. 
Extensive experiments demonstrate the superiority and robustness of EBMC. 
Ablation studies and further analyses confirm that enhancing weak modalities and dynamically adjusting modality contributions significantly improve model performance. 
Moreover, EBMC offers a unified perspective that bridges representation enhancement and modality coordination, yielding consistent gains across diverse emotion recognition tasks.

\section*{Acknowledgments}
This work is supported by the National Natural Science Foundation of China (No. 62176187).
We sincerely thank the authors of EMOE~\cite{fang2025emoe}, whose work inspired the visual design of several figures in this paper.

{
    \small
    \bibliographystyle{ieeenat_fullname}
    \bibliography{main}

@String(ICASSP=	{ICASSP})

@String(ICIP = {IEEE Int. Conf. Image Process.})

@String(AAAI = {AAAI})

@String(ICIP  = {ICIP})

@article{zadeh2016mosi,
  title={Mosi: multimodal corpus of sentiment intensity and subjectivity analysis in online opinion videos},
  author={Zadeh, Amir and Zellers, Rowan and Pincus, Eli and Morency, Louis-Philippe},
  journal={arXiv preprint arXiv:1606.06259},
  year={2016}
}

@inproceedings{bagher-zadeh-etal-2018-multimodal,
    title = "Multimodal Language Analysis in the Wild: {CMU}-{MOSEI} Dataset and Interpretable Dynamic Fusion Graph",
    author = "Bagher Zadeh, AmirAli  and
      Liang, Paul Pu  and
      Poria, Soujanya  and
      Cambria, Erik  and
      Morency, Louis-Philippe",
    booktitle = "Proceedings of the 56th Annual Meeting of the Association for Computational Linguistics (Volume 1: Long Papers)",
    year = "2018",
    pages = "2236--2246",
}

@article{busso2008iemocap,
  title={IEMOCAP: Interactive emotional dyadic motion capture database},
  author={Busso, Carlos and Bulut, Murtaza and Lee, Chi-Chun and Kazemzadeh, Abe and Mower, Emily and Kim, Samuel and Chang, Jeannette N and Lee, Sungbok and Narayanan, Shrikanth S},
  journal={Language resources and evaluation},
  volume={42},
  pages={335--359},
  year={2008},
}

@inproceedings{devlin2019bert,
  title={Bert: Pre-training of deep bidirectional transformers for language understanding},
  author={Devlin, Jacob and Chang, Ming-Wei and Lee, Kenton and Toutanova, Kristina},
  booktitle={Proceedings of the 2019 conference of the North American chapter of the association for computational linguistics: human language technologies, volume 1 (long and short papers)},
  pages={4171--4186},
  year={2019}
}

@article{maaten2008visualizing,
  title={Visualizing data using t-SNE},
  author={Maaten, Laurens van der and Hinton, Geoffrey},
  journal={Journal of machine learning research},
  volume={9},
  pages={2579--2605},
  year={2008}
}

@inproceedings{pennington-etal-2014-glove,
    title = "{G}lo{V}e: Global Vectors for Word Representation",
    author = "Pennington, Jeffrey  and
      Socher, Richard  and
      Manning, Christopher",
    booktitle = "Proceedings of the 2014 Conference on Empirical Methods in Natural Language Processing ({EMNLP})",
    year = "2014",
    pages = "1532--1543"
}

@inproceedings{degottex2014covarep,
  title={COVAREP—A collaborative voice analysis repository for speech technologies},
  author={Degottex, Gilles and Kane, John and Drugman, Thomas and Raitio, Tuomo and Scherer, Stefan},
  booktitle={2014 ieee international conference on acoustics, speech and signal processing (icassp)},
  pages={960--964},
  year={2014},
}

@inproceedings{wang2025dlf,
  title={DLF: Disentangled-language-focused multimodal sentiment analysis},
  author={Wang, Pan and Zhou, Qiang and Wu, Yawen and Chen, Tianlong and Hu, Jingtong},
  booktitle={Proceedings of the AAAI Conference on Artificial Intelligence},
  pages={21180--21188},
  year={2025}
}

@inproceedings{wang-etal-2024-refining,
    title = "Refining and Synthesis: A Simple yet Effective Data Augmentation Framework for Cross-Domain Aspect-based Sentiment Analysis",
    author = "Wang, Haining  and
      He, Kang  and
      Li, Bobo  and
      Chen, Lei  and
      Li, Fei  and
      Han, Xu  and
      Teng, Chong  and
      Ji, Donghong",
    booktitle = "Findings of the Association for Computational Linguistics: ACL 2024",
    year = "2024",
    pages = "10318--10329",
}

@inproceedings{hazarika2020misa,
  title={Misa: Modality-invariant and-specific representations for multimodal sentiment analysis},
  author={Hazarika, Devamanyu and Zimmermann, Roger and Poria, Soujanya},
  booktitle={Proceedings of the 28th Association for Computing Machinery International Conference on Multimedia},
  pages={1122--1131},
  year={2020}
}

@inproceedings{li2024unified,
  title={A unified self-distillation framework for multimodal sentiment analysis with uncertain missing modalities},
  author={Li, Mingcheng and Yang, Dingkang and Lei, Yuxuan and Wang, Shunli and Wang, Shuaibing and Su, Liuzhen and Yang, Kun and Wang, Yuzheng and Sun, Mingyang and Zhang, Lihua},
  booktitle={Proceedings of the AAAI conference on artificial intelligence},
  pages={10074--10082},
  year={2024}
}

@inproceedings{li2023decoupled,
  title={Decoupled multimodal distilling for emotion recognition},
  author={Li, Yong and Wang, Yuanzhi and Cui, Zhen},
  booktitle={Proceedings of the IEEE/CVF Conference on Computer vision and Pattern Recognition},
  pages={6631--6640},
  year={2023}
}

@inproceedings{yang2023confede,
  title={Confede: Contrastive feature decomposition for multimodal sentiment analysis},
  author={Yang, Jiuding and Yu, Yakun and Niu, Di and Guo, Weidong and Xu, Yu},
  booktitle={Proceedings of the 61st Annual Meeting of the Association for Computational Linguistics (Volume 1: Long Papers)},
  pages={7617--7630},
  year={2023}
}

@inproceedings{fang2025emoe,
  title={Emoe: Modality-specific enhanced dynamic emotion experts},
  author={Fang, Yiyang and Huang, Wenke and Wan, Guancheng and Su, Kehua and Ye, Mang},
  booktitle={Proceedings of the Computer Vision and Pattern Recognition Conference},
  pages={14314--14324},
  year={2025}
}

@inproceedings{fang2025catch,
  title={Catch your emotion: Sharpening emotion perception in multimodal large language models},
  author={Fang, Yiyang and Liang, Jian and Huang, Wenke and Li, He and Su, Kehua and Ye, Mang},
  booktitle={Forty-second International Conference on Machine Learning},
  year={2025}
}

@inproceedings{praveen2024recursive,
  title={Recursive joint cross-modal attention for multimodal fusion in dimensional emotion recognition},
  author={Praveen, R Gnana and Alam, Jahangir},
  booktitle={Proceedings of the IEEE/CVF Conference on Computer Vision and Pattern Recognition},
  pages={4803--4813},
  year={2024}
}

@inproceedings{shi2023multiemo,
  title={MultiEMO: An attention-based correlation-aware multimodal fusion framework for emotion recognition in conversations},
  author={Shi, Tao and Huang, Shao-Lun},
  booktitle={Proceedings of the 61st Annual Meeting of the Association for Computational Linguistics (Volume 1: Long Papers)},
  pages={14752--14766},
  year={2023}
}

@article{chen2025cr,
  title={CR-GAC: Cross-modal Recombination via Graph-Attention Collaborative Optimization for Multimodal Sentiment Analysis},
  author={Chen, Haoran and Liu, Jiapeng and Li, Zuhe and Pan, Yushan and Tao, Hongwei and Wu, Huaiguang and Wang, Yunyang and Yang, Chenguang},
  journal={Expert Systems with Applications},
  pages={129805},
  year={2025},
  publisher={Elsevier}
}

@article{he2025pase,
  title={PaSE: Prototype-aligned Calibration and Shapley-based Equilibrium for Multimodal Sentiment Analysis},
  author={He, Kang and Chen, Boyu and Ding, Yuzhe and Li, Fei and Teng, Chong and Ji, Donghong},
  journal={arXiv preprint arXiv:2511.17585},
  year={2025}
}

@inproceedings{he-etal-2025-dalr,
    title = "{DALR}: Dual-level Alignment Learning for Multimodal Sentence Representation Learning",
    author = "He, Kang  and
      Ding, Yuzhe  and
      Wang, Haining  and
      Li, Fei  and
      Teng, Chong  and
      Ji, Donghong",
    booktitle = "Findings of the Association for Computational Linguistics: ACL 2025",
    year = "2025",
    pages = "3586--3601",
}

@inproceedings{ding-etal-2025-zero,
    title = "Zero-Shot Conversational Stance Detection: Dataset and Approaches",
    author = "Ding, Yuzhe  and
      He, Kang  and
      Li, Bobo  and
      Zheng, Li  and
      He, Haijun  and
      Li, Fei  and
      Teng, Chong  and
      Ji, Donghong",
    booktitle = "Findings of the Association for Computational Linguistics: ACL 2025",
    year = "2025",
    pages = "3221--3235",
}

@inproceedings{li2023revisiting,
  title={Revisiting disentanglement and fusion on modality and context in conversational multimodal emotion recognition},
  author={Li, Bobo and Fei, Hao and Liao, Lizi and Zhao, Yu and Teng, Chong and Chua, Tat-Seng and Ji, Donghong and Li, Fei},
  booktitle={Proceedings of the 31st ACM International Conference on Multimedia},
  pages={5923--5934},
  year={2023}
}

@inproceedings{xu2025towards,
  title={Towards Multimodal Sentiment Analysis via Hierarchical Correlation Modeling with Semantic Distribution Constraints},
  author={Xu, Qinfu and Wei, Yiwei and Wu, Chunlei and Wang, Leiquan and Yuan, Shaozu and Wu, Jie and Lu, Jing and Zhou, Hengyang},
  booktitle={Proceedings of the AAAI Conference on Artificial Intelligence},
  pages={21788--21796},
  year={2025}
}

@inproceedings{zhang2025ecerc,
  title={ECERC: evidence-cause attention network for multi-modal emotion recognition in conversation},
  author={Zhang, Tao and Tan, Zhenhua},
  booktitle={Proceedings of the 63rd Annual Meeting of the Association for Computational Linguistics (Volume 1: Long Papers)},
  pages={2064--2077},
  year={2025}
}

@inproceedings{li2025alignmamba,
  title={AlignMamba: Enhancing Multimodal Mamba with Local and Global Cross-modal Alignment},
  author={Li, Yan and Xing, Yifei and Lan, Xiangyuan and Li, Xin and Chen, Haifeng and Jiang, Dongmei},
  booktitle={Proceedings of the Computer Vision and Pattern Recognition Conference},
  pages={24774--24784},
  year={2025}
}

@inproceedings{maniparambil2025harnessing,
  title={Harnessing Frozen Unimodal Encoders for Flexible Multimodal Alignment},
  author={Maniparambil, Mayug and Akshulakov, Raiymbek and Djilali, Yasser Abdelaziz Dahou and Narayan, Sanath and Singh, Ankit and O'Connor, Noel E},
  booktitle={Proceedings of the Computer Vision and Pattern Recognition Conference},
  pages={29847--29857},
  year={2025}
}

@inproceedings{das2023revisiting,
  title={Revisiting modality imbalance in multimodal pedestrian detection},
  author={Das, Arindam and Das, Sudip and Sistu, Ganesh and Horgan, Jonathan and Bhattacharya, Ujjwal and Jones, Edward and Glavin, Martin and Eising, Ciar{\'a}n},
  booktitle={2023 IEEE International Conference on Image Processing (ICIP)},
  pages={1755--1759},
  year={2023},
}

@inproceedings{wang2020makes,
  title={What makes training multi-modal classification networks hard?},
  author={Wang, Weiyao and Tran, Du and Feiszli, Matt},
  booktitle={Proceedings of the IEEE/CVF conference on computer vision and pattern recognition},
  pages={12695--12705},
  year={2020}
}

@inproceedings{wu2022characterizing,
  title={Characterizing and overcoming the greedy nature of learning in multi-modal deep neural networks},
  author={Wu, Nan and Jastrzebski, Stanislaw and Cho, Kyunghyun and Geras, Krzysztof J},
  booktitle={International Conference on Machine Learning},
  pages={24043--24055},
  year={2022},
}

@inproceedings{peng2022balanced,
  title={Balanced multimodal learning via on-the-fly gradient modulation},
  author={Peng, Xiaokang and Wei, Yake and Deng, Andong and Wang, Dong and Hu, Di},
  booktitle={Proceedings of the IEEE/CVF conference on computer vision and pattern recognition},
  pages={8238--8247},
  year={2022}
}

@inproceedings{huang2025adaptive,
  title={Adaptive unimodal regulation for balanced multimodal information acquisition},
  author={Huang, Chengxiang and Wei, Yake and Yang, Zequn and Hu, Di},
  booktitle={Proceedings of the Computer Vision and Pattern Recognition Conference},
  pages={25854--25863},
  year={2025}
}

@inproceedings{fan2024detached,
  title={Detached and interactive multimodal learning},
  author={Fan, Yunfeng and Xu, Wenchao and Wang, Haozhao and Liu, Junhong and Guo, Song},
  booktitle={Proceedings of the 32nd ACM International Conference on Multimedia},
  pages={5470--5478},
  year={2024}
}

@inproceedings{fan2023pmr,
  title={Pmr: Prototypical modal rebalance for multimodal learning},
  author={Fan, Yunfeng and Xu, Wenchao and Wang, Haozhao and Wang, Junxiao and Guo, Song},
  booktitle={Proceedings of the IEEE/CVF Conference on Computer Vision and Pattern Recognition},
  pages={20029--20038},
  year={2023}
}

@inproceedings{chen2025hyperdimensional,
  title={Hyperdimensional uncertainty quantification for multimodal uncertainty fusion in autonomous vehicles perception},
  author={Chen, Luke and Wang, Junyao and Mortlock, Trier and Khargonekar, Pramod and Al Faruque, Mohammad Abdullah},
  booktitle={Proceedings of the Computer Vision and Pattern Recognition Conference},
  pages={22306--22316},
  year={2025}
}

@inproceedings{gao2024embracing,
  title={Embracing unimodal aleatoric uncertainty for robust multimodal fusion},
  author={Gao, Zixian and Jiang, Xun and Xu, Xing and Shen, Fumin and Li, Yujie and Shen, Heng Tao},
  booktitle={Proceedings of the IEEE/CVF conference on computer vision and pattern recognition},
  pages={26876--26885},
  year={2024}
}

@inproceedings{li2025dpu,
  title={Dpu: Dynamic prototype updating for multimodal out-of-distribution detection},
  author={Li, Shawn and Gong, Huixian and Dong, Hao and Yang, Tiankai and Tu, Zhengzhong and Zhao, Yue},
  booktitle={Proceedings of the Computer Vision and Pattern Recognition Conference},
  pages={10193--10202},
  year={2025}
}

@inproceedings{zhang2025modal,
  title={Modal Feature Optimization Network with Prompt for Multimodal Sentiment Analysis},
  author={Zhang, Xiangmin and Wei, Wei and Zou, Shihao},
  booktitle={Proceedings of the 31st International Conference on Computational Linguistics},
  pages={4611--4621},
  year={2025}
}

@inproceedings{lee2023multimodal,
  title={Multimodal prompting with missing modalities for visual recognition},
  author={Lee, Yi-Lun and Tsai, Yi-Hsuan and Chiu, Wei-Chen and Lee, Chen-Yu},
  booktitle={Proceedings of the IEEE/CVF Conference on Computer Vision and Pattern Recognition},
  pages={14943--14952},
  year={2023}
}

@inproceedings{weng2025enhancing,
  title={Enhancing Multimodal Sentiment Analysis for Missing Modality through Self-Distillation and Unified Modality Cross-Attention},
  author={Weng, Yuzhe and Wang, Haotian and Gao, Tian and Li, Kewei and Niu, Shutong and Du, Jun},
  booktitle={ICASSP 2025-2025 IEEE International Conference on Acoustics, Speech and Signal Processing (ICASSP)},
  pages={1--5},
  year={2025},
}

@inproceedings{ma2023robust,
  title={Robust multiview multimodal driver monitoring system using masked multi-head self-attention},
  author={Ma, Yiming and Sanchez, Victor and Nikan, Soodeh and Upadhyay, Devesh and Atote, Bhushan and Guha, Tanaya},
  booktitle={Proceedings of the IEEE/CVF Conference on Computer Vision and Pattern Recognition},
  pages={2617--2625},
  year={2023}
}

@inproceedings{tao2025multi,
  title={A Multi-Focus-Driven Multi-Branch Network for Robust Multimodal Sentiment Analysis},
  author={Tao, Chuanqi and Li, Jiaming and Zang, Tianzi and Gao, Peng},
  booktitle={Proceedings of the AAAI Conference on Artificial Intelligence},
  pages={1547--1555},
  year={2025}
}

@inproceedings{tsai2025enhance,
  title={Enhance modality robustness in text-centric multimodal alignment with adversarial prompting},
  author={Tsai, Yun-Da and Yen, Ting-Yu and Liao, Keng-Te and Lin, Shou-De},
  booktitle={Proceedings of the AAAI Conference on Artificial Intelligence},
  pages={27740--27747},
  year={2025}
}

@inproceedings{tsai2019multimodal,
  title={Multimodal transformer for unaligned multimodal language sequences},
  author={Tsai, Yao-Hung Hubert and Bai, Shaojie and Liang, Paul Pu and Kolter, J Zico and Morency, Louis-Philippe and Salakhutdinov, Ruslan},
  booktitle={Proceedings of the conference. Association for computational linguistics. Meeting},
  pages={6558},
  year={2019}
}

@inproceedings{yu2021learning,
  title={Learning modality-specific representations with self-supervised multi-task learning for multimodal sentiment analysis},
  author={Yu, Wenmeng and Xu, Hua and Yuan, Ziqi and Wu, Jiele},
  booktitle={Proceedings of the AAAI conference on artificial intelligence},
  pages={10790--10797},
  year={2021}
}

@inproceedings{yu2023conki,
  title={ConKI: Contrastive Knowledge Injection for Multimodal Sentiment Analysis},
  author={Yu, Yakun and Zhao, Mingjun and Qi, Shi-ang and Sun, Feiran and Wang, Baoxun and Guo, Weidong and Wang, Xiaoli and Yang, Lei and Niu, Di},
  booktitle={Findings of the Association for Computational Linguistics: ACL 2023},
  pages={13610--13624},
  year={2023}
}

@inproceedings{yang2024clgsi,
  title={CLGSI: a multimodal sentiment analysis framework based on contrastive learning guided by sentiment intensity},
  author={Yang, Yang and Dong, Xunde and Qiang, Yupeng},
  booktitle={Findings of the Association for Computational Linguistics: NAACL 2024},
  pages={2099--2110},
  year={2024}
}

@inproceedings{wu2025enriching,
  title={Enriching multimodal sentiment analysis through textual emotional descriptions of visual-audio content},
  author={Wu, Sheng and He, Dongxiao and Wang, Xiaobao and Wang, Longbiao and Dang, Jianwu},
  booktitle={Proceedings of the AAAI Conference on Artificial Intelligence},
  pages={1601--1609},
  year={2025}
}

@inproceedings{gao2024enhanced,
  title={Enhanced Experts with Uncertainty-Aware Routing for Multimodal Sentiment Analysis},
  author={Gao, Zixian and Hu, Disen and Jiang, Xun and Lu, Huimin and Shen, Heng Tao and Xu, Xing},
  booktitle={Proceedings of the 32nd ACM International Conference on Multimedia},
  pages={9650--9659},
  year={2024}
}

@inproceedings{zhuang2024glomo,
  title={GLoMo: Global-Local Modal Fusion for Multimodal Sentiment Analysis},
  author={Zhuang, Yan and Zhang, Yanru and Hu, Zheng and Zhang, Xiaoyue and Deng, Jiawen and Ren, Fuji},
  booktitle={Proceedings of the 32nd ACM International Conference on Multimedia},
  pages={1800--1809},
  year={2024}
}

@inproceedings{lin2025semi,
  title={Semi-IIN: Semi-supervised Intra-inter modal Interaction Learning Network for Multimodal Sentiment Analysis},
  author={Lin, Jinhao and Wang, Yifei and Xu, Yanwu and Liu, Qi},
  booktitle={Proceedings of the AAAI Conference on Artificial Intelligence},
  pages={1411--1419},
  year={2025}
}

@inproceedings{pham2019found,
  title={Found in translation: Learning robust joint representations by cyclic translations between modalities},
  author={Pham, Hai and Liang, Paul Pu and Manzini, Thomas and Morency, Louis-Philippe and P{\'o}czos, Barnab{\'a}s},
  booktitle={Proceedings of the AAAI conference on artificial intelligence},
  pages={6892--6899},
  year={2019}
}

@inproceedings{wang2020transmodality,
  title={Transmodality: An end2end fusion method with transformer for multimodal sentiment analysis},
  author={Wang, Zilong and Wan, Zhaohong and Wan, Xiaojun},
  booktitle={Proceedings of the web conference 2020},
  pages={2514--2520},
  year={2020}
}

@inproceedings{ma2021smil,
  title={Smil: Multimodal learning with severely missing modality},
  author={Ma, Mengmeng and Ren, Jian and Zhao, Long and Tulyakov, Sergey and Wu, Cathy and Peng, Xi},
  booktitle={Proceedings of the AAAI Conference on Artificial Intelligence},
  pages={2302--2310},
  year={2021}
}

@article{lian2023gcnet,
  title={Gcnet: Graph completion network for incomplete multimodal learning in conversation},
  author={Lian, Zheng and Chen, Lan and Sun, Licai and Liu, Bin and Tao, Jianhua},
  journal={IEEE Transactions on pattern analysis and machine intelligence},
  volume={45},
  number={7},
  pages={8419--8432},
  year={2023},
  publisher={IEEE}
}

@inproceedings{li2024correlation,
  title={Correlation-decoupled knowledge distillation for multimodal sentiment analysis with incomplete modalities},
  author={Li, Mingcheng and Yang, Dingkang and Zhao, Xiao and Wang, Shuaibing and Wang, Yan and Yang, Kun and Sun, Mingyang and Kou, Dongliang and Qian, Ziyun and Zhang, Lihua},
  booktitle={Proceedings of the IEEE/CVF Conference on Computer Vision and Pattern Recognition},
  pages={12458--12468},
  year={2024}
}

@inproceedings{sun2022cubemlp,
  title={CubeMLP: An MLP-based model for multimodal sentiment analysis and depression estimation},
  author={Sun, Hao and Wang, Hongyi and Liu, Jiaqing and Chen, Yen-Wei and Lin, Lanfen},
  booktitle={Proceedings of the 30th ACM international conference on multimedia},
  pages={3722--3729},
  year={2022}
}

@article{wang2022cross,
  title={Cross-modal enhancement network for multimodal sentiment analysis},
  author={Wang, Di and Liu, Shuai and Wang, Quan and Tian, Yumin and He, Lihuo and Gao, Xinbo},
  journal={IEEE Transactions on Multimedia},
  volume={25},
  pages={4909--4921},
  year={2022},
  publisher={IEEE}
}

@article{han2021improving,
  title={Improving multimodal fusion with hierarchical mutual information maximization for multimodal sentiment analysis},
  author={Han, Wei and Chen, Hui and Poria, Soujanya},
  journal={arXiv preprint arXiv:2109.00412},
  year={2021}
}

@article{wang2023tetfn,
  title={TETFN: A text enhanced transformer fusion network for multimodal sentiment analysis},
  author={Wang, Di and Guo, Xutong and Tian, Yumin and Liu, Jinhui and He, LiHuo and Luo, Xuemei},
  journal={Pattern Recognition},
  volume={136},
  pages={109259},
  year={2023},
  publisher={Elsevier}
}

@inproceedings{poria2016convolutional,
  title={Convolutional MKL based multimodal emotion recognition and sentiment analysis},
  author={Poria, Soujanya and Chaturvedi, Iti and Cambria, Erik and Hussain, Amir},
  booktitle={2016 IEEE 16th international conference on data mining (ICDM)},
  pages={439--448},
  year={2016},
}

@inproceedings{yu2019adapting,
  title={Adapting BERT for target-oriented multimodal sentiment classification},
  author={Yu, Jianfei and Jiang, Jing},
  year={2019},
  booktitle={Proceedings of the 28th International Joint Conference on Artificial Intelligence}
}

@inproceedings{yuan2021transformer,
  title={Transformer-based feature reconstruction network for robust multimodal sentiment analysis},
  author={Yuan, Ziqi and Li, Wei and Xu, Hua and Yu, Wenmeng},
  booktitle={Proceedings of the 29th ACM international conference on multimedia},
  pages={4400--4407},
  year={2021}
}

@article{zhang2023learning,
  title={Learning language-guided adaptive hyper-modality representation for multimodal sentiment analysis},
  author={Zhang, Haoyu and Wang, Yu and Yin, Guanghao and Liu, Kejun and Liu, Yuanyuan and Yu, Tianshu},
  journal={arXiv preprint arXiv:2310.05804},
  year={2023}
}

@article{zhang2024towards,
  title={Towards robust multimodal sentiment analysis with incomplete data},
  author={Zhang, Haoyu and Wang, Wenbin and Yu, Tianshu},
  journal={Advances in Neural Information Processing Systems},
  volume={37},
  pages={55943--55974},
  year={2024}
}

@inproceedings{wang2019words,
  title={Words can shift: Dynamically adjusting word representations using nonverbal behaviors},
  author={Wang, Yansen and Shen, Ying and Liu, Zhun and Liang, Paul Pu and Zadeh, Amir and Morency, Louis-Philippe},
  booktitle={Proceedings of the AAAI conference on artificial intelligence},
  volume={33},
  number={01},
  pages={7216--7223},
  year={2019}
}

@inproceedings{zhai-etal-2024-chinese,
    title = "{C}hinese {M}ental{BERT}: Domain-Adaptive Pre-training on Social Media for {C}hinese Mental Health Text Analysis",
    author = "Zhai, Wei  and
      Qi, Hongzhi  and
      Zhao, Qing  and
      Li, Jianqiang  and
      Wang, Ziqi  and
      Wang, Han  and
      Yang, Bing  and
      Fu, Guanghui",
    booktitle = "Findings of the Association for Computational Linguistics: ACL 2024",
    year = "2024",
    pages = "10574--10585",
}

@inproceedings{zhang-etal-2024-escot,
    title = "{ESC}o{T}: Towards Interpretable Emotional Support Dialogue Systems",
    author = "Zhang, Tenggan  and
      Zhang, Xinjie  and
      Zhao, Jinming  and
      Zhou, Li  and
      Jin, Qin",
    booktitle = "Proceedings of the 62nd Annual Meeting of the Association for Computational Linguistics (Volume 1: Long Papers)",
    year = "2024",
    pages = "13395--13412",
}

@article{fernando2021missing,
 title={Missing the missing values: The ugly duckling of fairness in machine learning},
 author={Fernando, Mart{\'\i}nez-Plumed and C{\`e}sar, Ferri and David, Nieves and Jos{\'e}, Hern{\'a}ndez-Orallo},
 journal={International Journal of Intelligent Systems},
 volume={36},
 number={7},
 pages={3217--3258},
 year={2021},
}

@article{oord2018representation,
  title={Representation learning with contrastive predictive coding},
  author={Oord, Aaron van den and Li, Yazhe and Vinyals, Oriol},
  journal={arXiv preprint arXiv:1807.03748},
  year={2018}
}
}

\appendix
\clearpage
\setcounter{page}{1}
\maketitlesupplementary

\section{Experimental Configuration}
\label{appendix:experments_configuration}

\subsection{Baselines}
\label{appendix:more_baselines}

In this study, to systematically evaluate model performance under full-modality, missing-modality, and cross-task scenarios, we adopt a comprehensive suite of state-of-the-art baselines covering multimodal fusion, semantic disentanglement, robustness modeling, and imbalance mitigation.
\begin{itemize}
    \item MulT \cite{tsai2019multimodal}: proposes a Multimodal Transformer that models unaligned multimodal sequences using directional cross-modal attention, enabling interactions across modalities and time steps while capturing long-range dependencies.
    \item SelfMM \cite{yu2021learning}: proposes a self-supervised framework that generates unimodal labels to enable joint training of multimodal and unimodal tasks, using a dynamic weight-adjustment strategy to balance subtasks and enhance the capture of modality-specific differences.
    \item ConKI \cite{yu2023conki}: proposes a framework that injects domain-specific and general knowledge via contrastive learning to enhance multimodal sentiment prediction.
    \item ConFEDE \cite{yang2023confede}: proposes a unified framework that enhances multimodal sentiment representations by jointly performing contrastive representation learning and feature decomposition, separating modality-invariant and modality-specific features across text, video, and audio.
    \item CLGSI \cite{yang2024clgsi}: proposes sentiment-intensity-guided contrastive learning with a fusion mechanism to jointly capture common and modality-specific features for multimodal sentiment prediction.
    \item DEVA \cite{wu2025enriching}: proposes a progressive fusion framework that converts audio-visual inputs into textualized sentiment descriptions and fuses them via a text-guided module to capture subtle emotional variations.
    \item EUAR \cite{gao2024enhanced}: proposes a Mixture-of-Experts framework with uncertainty-aware routing to handle noisy data in MSA, dynamically directing samples to experts with lower uncertainty to extract clearer features.
    \item EMOE \cite{fang2025emoe}: proposes a Mixture-of-Modality-Experts framework that dynamically weights modalities per sample via a router network, and uses unimodal distillation to retain single-modality predictive ability within fused features. We reproduce it using utterance-level features to align with our evaluation protocol.
    \item GLoMo \cite{zhuang2024glomo}: proposes a Global-Local Fusion framework that integrates local representations via modality-specific experts and combines them with global features through a guided fusion module, enhancing multimodal sentiment, humor, and emotion analysis.
    \item Semi-IIN \cite{lin2025semi}: proposes a semi-supervised network that captures intra- and inter-modal interactions using masked attention and gating, dynamically selects important features, and leverages unlabeled data to improve MSA.
    \item MCTN \cite{pham2019found}: learns joint representations via modality translation with cycle-consistency, enabling robust sentiment prediction from a single modality.
    \item TransM \cite{wang2020transmodality}: proposes a Transformer-based fusion method that translates between modalities to learn joint representations, improving MSA.
    \item SMIL \cite{ma2021smil}: proposes a Bayesian meta-learning framework for multimodal learning with missing modalities during training and testing, achieving robust performance even with severely incomplete data.
    \item GCNet \cite{lian2023gcnet}: proposes a graph-based framework for incomplete multimodal conversation understanding, using Speaker and Temporal GNNs to capture dependencies and jointly optimizing classification and reconstruction.
    \item UMDF \cite{li2024unified}: proposes a self-distillation framework with multi-grained crossmodal interaction and dynamic feature integration to handle uncertain missing modalities, producing robust multimodal representations for sentiment analysis.
    \item DMD \cite{li2023decoupled}: proposes Decoupled Multimodal Distillation that separates each modality into modality-irrelevant and modality-exclusive spaces and applies a dynamic graph distillation unit to flexibly transfer crossmodal knowledge, enhancing emotion recognition.
    \item CorrKD \cite{li2024correlation}: proposes a knowledge distillation framework for uncertain missing modalities, using contrastive, prototype-guided, and response-disentangled strategies to reconstruct missing semantics and improve MSA.
    \item LNLN \cite{zhang2024towards}: proposes language-dominated framework with dominant-modality correction and multimodal learning modules to enhance robustness and performance in MSA under noisy or missing data.
    
\end{itemize}

\subsection{Implementation Details}
\label{appendix:more_implementation_details}
To ensure fair and reproducible comparisons with prior work, we follow the standard feature extraction settings widely adopted in multimodal sentiment/emotion analysis.

\vspace{-4mm}
\paragraph{Textual Modality.}
We extract textual features using the \texttt{BERT$_\textrm{base}$} model~\cite{devlin2019bert}.
Each utterance is encoded through the 12-layer Transformer encoder, from which we take the contextualized hidden states of size 768 as the word-level representations.
Following established IEMOCAP ERC protocols, we additionally incorporate 300-dimensional GloVe embeddings~\cite{pennington-etal-2014-glove}.

\vspace{-4mm}
\paragraph{Visual Modality.}
We adopt the Facet toolkit to derive a 35-dimensional vector of facial action unit (AU) features for each frame.
The AU descriptors capture fine-grained facial muscle activations related to emotional expressions.
For utterance-level features, the AU sequences are first temporally aligned with the text and audio streams and then aggregated using mean pooling.

\vspace{-4mm}
\paragraph{Acoustic Modality.}
We use the COVAREP toolkit~\cite{degottex2014covarep} to extract 74-dimensional low-level descriptors, including glottal source parameters, prosodic cues, spectral features, and other physiologically interpretable features.
Frame-level features are resampled to match the temporal resolution of other modalities before being aggregated into utterance-level representations.

\vspace{-4mm}
\paragraph{Training Setup.}
All hyperparameters in EBMC are tuned on the validation set.
Unless otherwise specified, we set $\lambda_1, \lambda_2, \beta, \gamma, \eta = 0.1$ and $\zeta = 0.5$, which provides a stable balance among reconstruction, semantic enhancement, and modality collaboration objectives.

Each stage of the EBMC framework is trained independently, and we tailor the batch size and training schedule to the characteristics of each dataset.
Unless otherwise specified, the learning rate is fixed at 0.0001, which we found to provide stable convergence across all datasets.
For CMU-MOSEI, due to its large scale and substantial modality diversity, we adopt a batch size of 32. The full model is trained for 200 epochs, with 100 epochs allocated to each stage of EBMC.
For CMU-MOSI, which is smaller but still rich in modality-specific variations, we use a batch size of 64 to accelerate training.
The model is trained for 300 epochs, with 150 epochs per stage, ensuring adequate optimization in both the enhancement and collaboration phases.
For IEMOCAP, given its limited size and higher speaker variability, we reduce the batch size to 16 to maintain stable gradient estimates.
Each stage of the framework is trained for 150 epochs, providing a balanced schedule that avoids overfitting while ensuring sufficient convergence.
All experiments are conducted on a workstation equipped with an NVIDIA RTX 4090 GPU (48GB).
Training is implemented in PyTorch with mixed-precision acceleration enabled by default.

\subsection{Computational Cost and Complexity Analysis}
\label{appendix:computational_cost_time_complexity}

The computational complexity of EBMC can be derived directly from the operations defined in MSD, CCE, EMC, and IMTD. 
Let $T_m$ and $d_m$ denote the temporal length and feature dimension of modality $m$, and let $h_m$ be the hidden width of the lightweight MLPs. 

In MSD, the shared–specific decomposition requires two forward passes through modality-dependent MLPs, resulting in a cost of $O(T_m d_m h_m)$. 
The invariant alignment term based on InfoNCE further introduces a complexity of $O(|M|\, T_m d_m)$ due to pairwise similarities and softmax normalization, while the decorrelation of modality-specific components incurs an additional $O(|M|^2 d_m)$ complexity through cosine-similarity computations. 
In CCE, the enhancement network $G_m$ operates on compact aggregated vectors and therefore contributes only $O(d_m h_m)$ per modality. 
EMC consists solely of vector norms, entropy terms, energy differences, and energy gradients such as $\|z_m\|_2^2$, $H(p_{T_m})$, and $(E(m_i)-E(m_j))^2$, each of which is computed in $O(d_m)$ time. 
IMTD computes variances, confidence scores, and KL divergence terms in $O(T_m d_m)$ time without introducing additional nonlinear transformations. 

Summing over all modalities yields the total computational complexity of EBMC:
\begin{equation}
O\!\left(
\sum_{m}
\left[
T_m d_m h_m
+
|M|\, T_m d_m
+
|M|^2 d_m
\right]
\right),
\end{equation}
which is dominated by the linear-time operations in MSD. Since EMC and IMTD consist exclusively of element-wise algebraic operations, they contribute negligible overhead relative to the backbone encoders and do not introduce attention-like quadratic costs at any stage of the framework.

\begin{table}[t]
\centering
\caption{Comparison of training efficiency and parameters across different baselines on CMU-MOSI dataset. Please see Appendix~\ref{appendix:training_efficiency} for details.}
\begin{tabular}{l|cc}
\toprule
\textbf{Model} & \textbf{Training Time} & \textbf{Params} \\
\midrule
SelfMM & 3.73h & 121,835,723 \\
GLoMo  & 2.29h & 129,818,887 \\
EUAR   & 1.16h & 110,436,422 \\
EBMC$_\textrm{online}$   & 1.04h & 112,320,746 \\
EBMC$_\textrm{offline}$ & 5min & 6,384,400 \\

\bottomrule
\end{tabular}

\label{tab:training_efficiency}
\vspace{-3mm}
\end{table}

\subsection{Empirical Training Efficiency and Parameter Comparison}
\label{appendix:training_efficiency}

In addition to the theoretical analysis in Appendix~\ref{appendix:computational_cost_time_complexity},
we further report the empirical training efficiency of EBMC and other baselines in \cref{tab:training_efficiency}.
All results are measured on the CMU-MOSI dataset by training each model for 300 epochs, ensuring a fair comparison across methods.

When EBMC is trained in the standard online setting—where raw inputs are loaded on-the-fly and each batch requires passing all modalities through BERT-based encoders—the end-to-end training takes approximately 1.2 hours, and the full system consists of 112,320,746 parameters, most of which originate from the Transformer encoders.
In contrast, if multimodal features are pre-extracted offline and cached on disk, the training pipeline no longer depends on the heavy BERT encoders. Under this offline-feature configuration, EBMC reduces to only the lightweight MSD, CCE, EMC, and IMTD modules, resulting in a compact model with merely 6,384,400 parameters, and the total training time for 300 epochs drops dramatically to around 5 minutes.

This comparison highlights that the majority of computational overhead in multimodal sentiment models stems from online text encoding, and that EBMC itself remains lightweight and efficient when decoupled from the feature extractor.

\section{Theoretical Foundations of Energy-guided Modality Coordination (EMC)}
\label{appendix:emc_theory}

This appendix provides a rigorous theoretical formulation of the proposed 
Energy-guided Modality Coordination module. 
We first establish its interpretation as a structured Energy-based Model (EBM), 
then analyze its connection and distinction from traditional gradient-balancing methods, 
followed by a dynamical perspective on the induced energy dynamics, 
and finally present the full gradients of the EMC objective.

\subsection{From Modality Energy to Energy-based Model}

For each modality $m$, the main paper defines a modality-specific energy potential:
\begin{equation}
E(m)= 
\alpha \| z_m \|_2^2 
+ \beta\, \ell_m
+ \gamma\, u_m ,
\label{eq:appendix_energy}
\end{equation}
where $z_m$ is the latent representation, $\ell_m$ is the modality-specific loss,
and $u_m$ denotes predictive uncertainty:
\begin{equation}
u_m 
= 
\mathbb{E}_i\!\left[ H\!\left(p_{T_m}^i(y)\right) \right], 
\qquad 
H(p) = - \sum_{y} p(y)\log p(y).
\end{equation}

\vspace{3pt}
\noindent\textbf{Joint Multimodal Energy.}
Let $M$ denote the set of modalities.
We define the global multimodal energy as:
\begin{equation}
E_{\text{joint}}
=
\sum_{m \in M} E(m)
+
\eta \sum_{m<m'} \big(E(m)-E(m')\big)^2 ,
\label{eq:joint_energy}
\end{equation}
where the second term encourages \emph{energy equilibrium} across modalities by penalizing pairwise energy gaps.

The joint distribution of all modality representations then forms a Gibbs distribution:
\begin{equation}
p_\theta(\{z_m\})
=
\frac{%
\exp\!\left(-E_{\text{joint}}(\{z_m\})\right)
}{
Z(\theta)
},
\label{eq:gibbs}
\end{equation}
which establishes EMC as a structured EBM over multimodal representations.

\vspace{3pt}
\noindent\textbf{Energy-descent Dynamics.}
Classical EBM learning often involves a Langevin-style update:
\begin{equation}
z^{t+1}=z^t - \lambda \,\nabla_z E_{\text{joint}}(z^t) + \omega^t,
\qquad 
\omega^t\sim \mathcal{N}(0,\lambda I).
\label{eq:langevin}
\end{equation}

EMC adopts the deterministic gradient-descent component at the modality level:
\begin{equation}
\Delta z_m
=
- \lambda 
\frac{\partial E(m)}{\partial z_m},
\label{eq:appendix_flow}
\end{equation}
which can be interpreted as a single-step Langevin gradient flow on the energy manifold, 
restricted to the modality-specific potential $E(m)$.
This establishes a principled EBM interpretation of the EMC training dynamics.

\subsection{Relation to Gradient-balancing Approaches}

\paragraph{Equivalence in Form.}
Traditional gradient-balancing methods modify per-modality gradients via
\begin{equation}
g_m^{\text{new}} = w_m g_m ,
\label{eq:gb_template}
\end{equation}
where $g_m$ is the original gradient and $w_m$ is an explicitly designed weight.
In contrast, EMC introduces an energy-gap loss: 
\begin{equation}
\mathcal{L}_{gap}
=
\sum_{m<m'}\big(E(m)-E(m')\big)^2 .
\end{equation}

Taking the derivative w.r.t.\ $z_m$ gives
\begin{equation}
\frac{\partial \mathcal{L}_{gap}}{\partial z_m}
=
2\sum_{m'\neq m}(E(m)-E(m'))\,
\frac{\partial E(m)}{\partial z_m}.
\label{eq:implicit_reweight}
\end{equation}

Letting $\bar E = \frac{1}{|M|}\sum_{k \in M} E(k)$, we observe
\begin{equation}
\sum_{m'\neq m}(E(m)-E(m'))
=
|M|\big(E(m)-\bar E\big),
\end{equation}
which implies an implicit, energy-aware weight of the form
\begin{equation}
w_m^{\text{EMC}}
\propto E(m)-\bar E.
\end{equation}
Hence EMC behaves as an \emph{energy-aware gradient modulator} whose weights are not manually chosen but emerge from the energy-gap objective.

\vspace{3pt}
\noindent\textbf{Fundamental Difference.}
Unlike explicit gradient-balancing heuristics,
EMC introduces a \emph{global energy potential} and optimizes
the system via its gradient flow.
Thus, modality correction emerges \emph{naturally} from the
energy equilibrium rather than manual gradient manipulation.

Formally, EMC optimizes
\begin{equation}
\min_{\{z_m\}}
\;\;
\mathcal{L}_{EMC}
=
\mathcal{L}_{gap}
+
\delta
\sum_{m\in M} 
\left\|
\nabla_{z_m} E(m)
\right\|^2,
\end{equation}
which is a well-defined and fully differentiable objective.
This grants EMC stronger theoretical grounding than 
ad-hoc gradient-balancing rules.

\begin{figure*}[t]
\centering \includegraphics[width=2.08 \columnwidth]{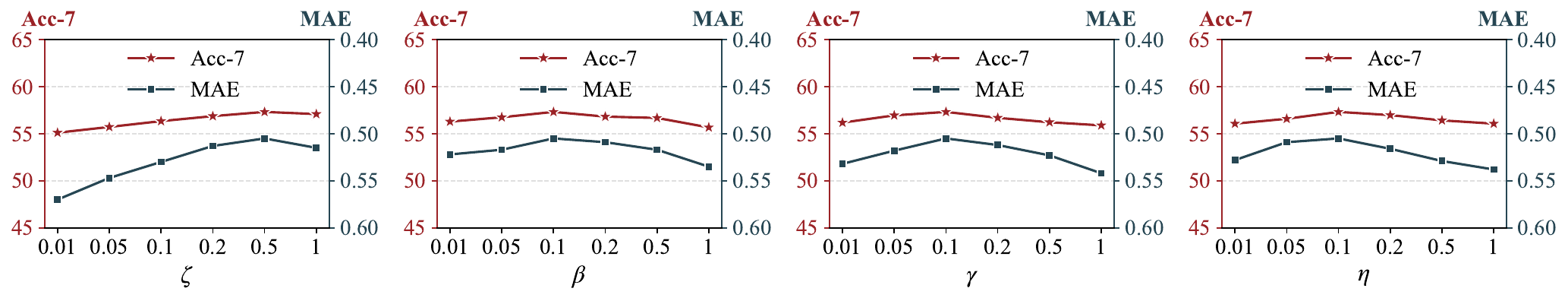}
\caption{
    Hyperparameter sensitivity analysis of EBMC on the CMU-MOSEI dataset. The effects of varying $\beta$, $\gamma$, $\eta$, and $\zeta$ are evaluated 
    according to the overall training objective defined in Eq.~\eqref{eq:total_loss}, while fixing the remaining hyperparameters to the values used in our main experiments. Please refer to Appendix \ref{appendix:hyperparameter_analysis}.}
    
\label{fig:appdenix_parameter_analysis}
\vspace{-3mm}
\end{figure*}

\subsection{Dynamical Analysis of EMC}

Let $g_m=\nabla_{z_m} E(m)$ and again denote $\bar E = \frac{1}{|M|}\sum_{k\in M}E(k)$.

\paragraph{Negative-feedback Structure.}
From \cref{eq:implicit_reweight}, the EMC-induced dynamics are proportional to
\begin{equation}
\Delta z_m
\propto
-\big(E(m)-\bar E\big)\, g_m ,
\label{eq:feedback}
\end{equation}
up to a constant factor.
This forms a classical negative-feedback system:

- if $E(m)>\bar E$ (weak or under-optimized modality), the factor $(E(m)-\bar E)$ is positive and the gradient magnitude is effectively increased $\rightarrow$ \emph{amplification};  

- if $E(m)<\bar E$ (dominant modality), the factor becomes negative and the effective gradient is reduced $\rightarrow$ \emph{suppression}.

Thus EMC automatically stabilizes modality interactions by amplifying high-energy modalities and suppressing low-energy ones.

\vspace{3pt}
\paragraph{Stability and Curvature Control.}
The regularization term 
$\delta\sum_m \|\nabla_{z_m}E(m)\|^2$ 
introduces an additional dependence on the curvature of $E(m)$.
Writing $H_m=\nabla_{z_m}^2 E(m)$ for the Hessian, 
the local second-order behavior of $\mathcal{L}_{EMC}$ along $z_m$
contains terms of the form
\begin{equation}
2\|g_m\|^2
\quad\text{and}\quad
2\big(E(m)-\bar E\big)\,H_m,
\end{equation}
up to higher-order derivatives.
Intuitively, the first term penalizes large gradients and thus smooths sharp changes in the energy landscape, 
while the second term reduces curvature as modalities approach equilibrium, 
contributing to stable optimization dynamics.

\paragraph{Convergence.}
A stationary point of EMC satisfies
\begin{equation}
E(m_1)=\cdots=E(m_{|M|}),
\qquad
\nabla_{z_m}E(m)=0,
\quad\forall m.
\end{equation}

At such points, all modalities lie in a balanced energy configuration
and each representation reaches a local energy minimum,
corresponding to a stable multimodal fusion state.

\subsection{Derivatives of EMC Components}

\paragraph{Energy Gradient.}
From \cref{eq:appendix_energy},
\begin{equation}
\nabla_{z_m}E(m)
=
2\alpha z_m
+
\beta \nabla_{z_m}\ell_m
+
\gamma \nabla_{z_m} u_m .
\end{equation}

\paragraph{Derivative of Predictive Uncertainty.}
For the entropy,
\begin{equation}
\frac{\partial H(p)}{\partial p(y)}
=
-\log p(y)-1.
\end{equation}
Thus:
\begin{equation}
\nabla_{z_m}u_m
=
\mathbb{E}_i
\left[
\sum_y
\big(-\log p_{T_m}^i(y)-1\big)
\,
\nabla_{z_m}p_{T_m}^i(y)
\right].
\end{equation}
If $p_{T_m}^i(y)$ is produced by softmax logits $s_m^i(y)$, we have
\begin{equation}
\frac{\partial p_{T_m}^i(y)}{\partial s_m^i(y')}
=
p_{T_m}^i(y)\big(\mathbb{I}_{y=y'} - p_{T_m}^i(y')\big).
\end{equation}

\paragraph{Gradient of the Full EMC Objective.}
The full EMC objective can be written as
\begin{equation}
\mathcal{L}_{EMC}
=
\sum_{m<m'}\big(E(m)-E(m')\big)^2
+
\delta\sum_{m\in M} 
\left\|
\nabla_{z_m}E(m)
\right\|^2.
\end{equation}
Differentiating w.r.t.\ $z_m$ yields
\begin{align}
\nabla_{z_m}\mathcal{L}_{EMC}
=&\;
2
\sum_{m'\neq m} 
\big(E(m)-E(m')\big)\,
\nabla_{z_m}E(m)
\nonumber\\
&+
2\delta\, H_m\, g_m,
\label{eq:appendix_final_grad}
\end{align}
where $g_m=\nabla_{z_m}E(m)$ and $H_m=\nabla_{z_m}^2E(m)$.
The first term enforces pairwise energy equilibrium,
while the second term regularizes the curvature of the energy landscape,
jointly ensuring stable multimodal coordination.

\begin{figure*}[ht!]
\centering \includegraphics[width=2.08 \columnwidth]{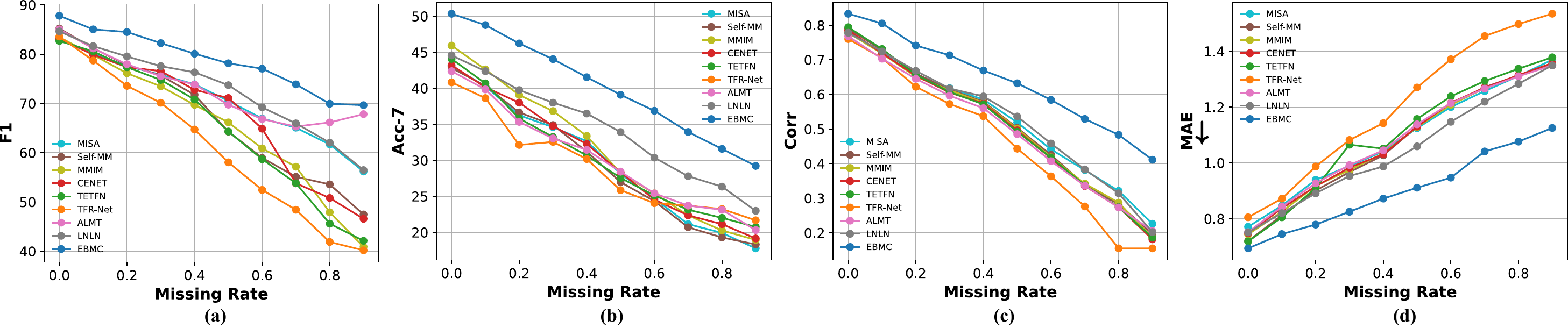}
\caption{
    Performance curves on the CMU-MOSI dataset under increasing modality missing rates. The four subplots respectively report F1, Acc-7, Corr, and MAE across all compared models. Note that F1, Acc-7, and Corr are \emph{higher-is-better} metrics, whereas MAE is a \emph{lower-is-better} metric. Please refer to \cref{sec:analyses_discussion} (Q3) and Appendix \ref{appendix:robust_comparison}.}
    
\label{fig:appdenix_curve_MOSI}

\end{figure*}
\begin{figure*}[ht!]
\centering \includegraphics[width=2.08 \columnwidth]{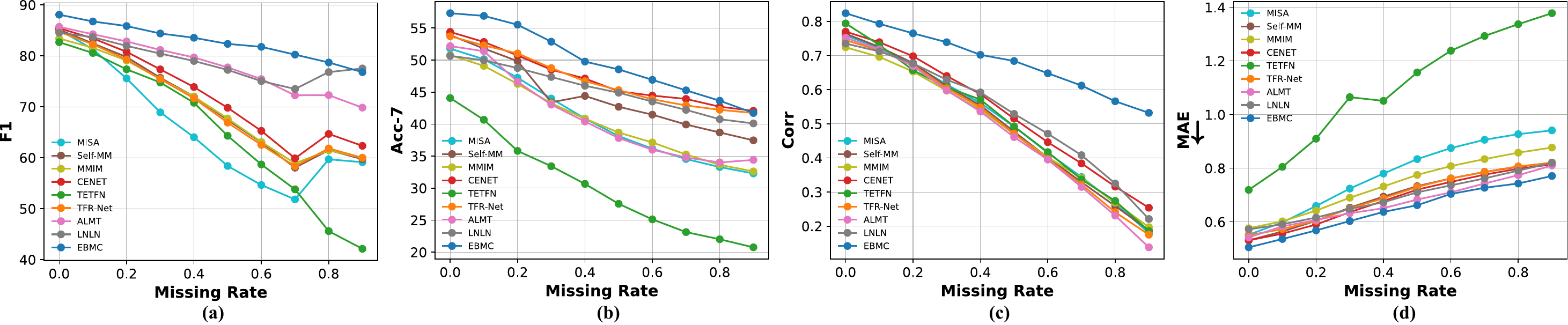}
\caption{
    Performance curves on the CMU-MOSEI dataset under increasing modality missing rates. The four subplots respectively report F1, Acc-7, Corr, and MAE across all compared models. Note that F1, Acc-7, and Corr are \emph{higher-is-better} metrics, whereas MAE is a \emph{lower-is-better} metric. Please refer to \cref{sec:analyses_discussion} (Q3) and Appendix \ref{appendix:robust_comparison}.}
    
\label{fig:appdenix_curve_MOSEI}
\vspace{-3mm}
\end{figure*}

\section{Additional Experiments and Analysis}
\label{appendix:more_experiments}

\subsection{Sensitivity analysis}
\label{appendix:hyperparameter_analysis}

We evaluate the sensitivity of EBMC to the four coefficients in
\cref{eq:total_loss}. Each coefficient is varied
independently while keeping the others fixed, and we report Acc-7 and MAE on
CMU-MOSEI.
As shown in~\cref{fig:appdenix_parameter_analysis}, the coefficients
$\beta$ (CCE), $\gamma$ (EMC), and $\eta$ (IMTD) all exhibit a consistent trend: performance peaks at the moderate value of $0.1$, achieving the best results.
This suggests that a balanced level of regularization for these three objectives is most effective, whereas both under- and over-regularization lead to performance drops.
The MSD coefficient $\zeta$ follows a different trend: performance improves
monotonically up to $\zeta=0.5$, where the same optimal score is achieved, and
declines slightly at $\zeta=1.0$. This indicates that MSD benefits from a
stronger weight, but excessive emphasis becomes counterproductive.
Overall, EBMC demonstrates stable behavior across a broad hyperparameter range, and the configuration used in the main experiments ($\beta=\gamma=\eta=0.1,\ \zeta=0.5$) closely matches the empirically optimal setting.

\subsection{Contribution of Energy-guided Modality Coordination}
\label{appendix:emc_ablation}
To better understand how each energy component contributes to EMC, we perform an ablation study by removing one term at a time (Table \ref{tab:emc_ablation}).
Across all three metrics (F1, Acc-7, and MAE), every variant that removes a single component performs worse than the full EBMC model, showing that each term contributes a unique and irreplaceable effect.

Removing $|z_m|_2^2$ leads to a noticeable decline (e.g., Acc-7 drops from 57.32 to 56.62), indicating that the quadratic penalty helps stabilize the latent space and prevents uncontrolled shifts in modality embeddings.
Eliminating the loss-coupling term $\ell_m$ results in the largest degradation (Acc-7=55.94), which aligns with its function of anchoring the energy landscape to task-relevant supervision.
Omitting the uncertainty term $u_m$ also weakens performance, demonstrating that entropy-based calibration is valuable for distinguishing reliable from unreliable modalities during energy balancing.
Overall, these results show that EMC is not a simple heuristic combination of losses; each component makes a substantive contribution to effective multimodal coordination.

\begin{table}[t]
\centering
\caption{Ablation of EMC energy components on CMU-MOSEI. We report F1 ($\uparrow$), Acc-7 ($\uparrow$), and MAE ($\downarrow$). Refer to Appendix \ref{appendix:emc_ablation}.}

\begin{adjustbox}{width=0.85\columnwidth}
\begin{tabular}{l|ccc}
\toprule
&  F1 ($\uparrow$) & Acc-7 ($\uparrow$) & MAE ($\downarrow$) \\
\midrule
\midrule
\rowcolor{Color1!80}

EBMC & \textbf{86.23/88.07} & \textbf{57.32} & \textbf{0.505} \\
\quad w/o $\|z_m\|_2^2$ 
& 85.54/87.52 & 56.62 & 0.518 \\
\rowcolor{Color1!80}
\quad w/o $\ell_m$ 
& 85.05/86.89 & 55.94 & 0.541 \\
\quad w/o $u_m$ 
& 85.36/87.16 & 56.40 & 0.527 \\
\bottomrule
\end{tabular}

\end{adjustbox}
\label{tab:emc_ablation}
\vspace{-3mm}
\end{table}

\subsection{Extended Robustness Comparison under Random Missing Rates}
\label{appendix:robust_comparison}

Across \cref{fig:appdenix_curve_MOSI} and \cref{fig:appdenix_curve_MOSEI}, we observe a clear performance contrast between EBMC and existing baselines as the missing rate increases. 
Most baselines degrade rapidly once modality inputs become unreliable, revealing their strong dependence on fully observed data. 
In contrast, EBMC exhibits a noticeably smoother and more stable degradation trend. This resilience is attributed to two principles: 
the disentanglement of shared and modality-specific factors, which prevents excessive reliance on any single modality, and the enhancement–then–balance mechanism, which maintains cross-modal semantic cues even when certain modalities fail. 
As a result, EBMC avoids the majority-class collapse commonly seen in other methods under high missing rates and retains discriminative capability across the full corruption spectrum.

Another consistent observation is that nearly all models, including EBMC, perform better on CMU-MOSEI than on CMU-MOSI under the same missing rate. 
This difference arises from MOSEI’s substantially larger scale and richer modality diversity, which provide stronger supervision and more stable temporal–semantic patterns. 
The additional data effectively buffers the impact of missing modalities, leading to improved robustness across models. 
EBMC benefits from these properties while still maintaining a clear advantage over competing approaches on both datasets, demonstrating that its robustness generalizes across data regimes rather than relying on dataset-specific artifacts.

\end{document}